\documentclass[preprint,12pt]{elsarticle}

\usepackage{natbib}
\usepackage{amssymb}
\usepackage[utf8]{inputenc}
\usepackage{graphicx}
\usepackage{xcolor}
\usepackage{tikz}
\usetikzlibrary{mindmap}
\usepackage{multirow}
\usepackage{amssymb}
\usepackage[shortlabels]{enumitem}
\usepackage{boldline} 
\usepackage{ulem}
\usepackage{smartdiagram}
\usepackage{tikz}
\usepackage{longtable}
\usepackage{hyperref}

\usesmartdiagramlibrary{additions}
\usetikzlibrary{fit,shapes.geometric,bending}

\makeatletter
\renewcommand\paragraph{\@startsection{paragraph}{4}{\z@}%
                                     {-3.25ex\@plus -1ex \@minus -.2ex}%
                                     {1.5ex \@plus .2ex}%
                                     {\normalfont\normalsize\itshape}}
\makeatother

\newcommand{\todo}[1]{\textbf{\color{red} #1}} 

\journal{Journal of Biomedical Informatics}

\begin{document}

\begin{frontmatter}
\title{MLHOps: Machine Learning for Healthcare Operations }

\author[label1]{Faiza Khan Khattak}
\author[label1,label2,label3]{Vallijah Subasri}
\author[label1]{Amrit Krishnan}
\author[label1]{Elham Dolatabadi}
\author[label1]{Deval Pandya}
\author[label4]{Laleh Seyyed-Kalantari}
\author[label1,label3,label5]{Frank Rudzicz}

\address[label1]{Vector Institute for Artificial Intelligence, Toronto, Ontario, Canada}
\address[label2]{Hospital for Sick Children, Toronto, Ontario, Canada} 
\address[label3]{University of Toronto, Toronto, Ontario, Canada}
\address[label4]{York University, Toronto, Ontario, Canada}
\address[label5]{Dalhousie University, Halifax, Nova Scotia, Canada}
\begin{abstract}
Machine Learning Health Operations (MLHOps) is the combination of processes for reliable, efficient, usable, and ethical deployment and maintenance of machine learning models in healthcare settings. This paper provides both a survey of work in this area and guidelines for developers and clinicians to deploy and maintain their own models in clinical practice.  We cover the foundational concepts of general machine learning operations, describe the initial setup of MLHOps pipelines (including data sources, preparation, engineering, and tools). We then describe long-term monitoring and updating (including data distribution shifts and model updating) and ethical considerations (including bias, fairness, interpretability, and privacy). This work therefore provides guidance across the full pipeline of MLHOps from conception to initial and ongoing deployment.
\end{abstract}

\end{frontmatter}

\section{Introduction}

Over the last decade, efforts to use health data for solving complex medical problems have increased significantly. Academic hospitals are increasingly dedicating resources to bring machine learning (ML) to the bedside and to addressing issues encountered by clinical staff. These resources are being utilized across a range of applications including clinical decision support, early warning, treatment recommendation, risk prediction, image informatics,  tele-diagnosis, drug discovery, and intelligent health knowledge systems. 

There are various examples of ML being applied to medical data, including prediction of sepsis \cite{sendakreal}, in-hospital mortality, prolonged length-of-stay, patient deterioration, and unplanned readmission \cite{rajkomar2018scalable}. In particular, sepsis is one of the leading causes of in-hospital deaths. A large-scale study demonstrated the impact of an early warning system to reduce the lead time for detecting the onset of sepsis, and hence allowing more time for clinicians to prescribe antibiotics \cite{adams2022prospective}. Similarly, deep convolutional neural networks have been shown to achieve superior performance in detecting pneumonia and other pathologies from chest X-rays, compared to practicing radiologists \cite{rajpurkar2017chexnet}. These results highlight the potential of ML models when they are strongly integrated into clinical workflows. \\

When deployed successfully, data-driven models can free time for clinicians\cite{henry2022human}, improve clinical outcomes \cite{rajkomar2019machine}, reduce costs \cite{bates2014big}, and provide improved quality care for patients. However, most studies remain preliminary, limited to small datasets, and/or implemented in select health sub-systems. Integrating with clinical workflows remains crucial \cite{wiens2019no,verma2021health} but, despite recent computational advances and an explosion of health data, deploying ML in healthcare responsibly and reliably faces several operational and engineering challenges, including: 

\begin{itemize}
    \item Standardizing data formats,
    \item Strengthening methodologies for evaluation, monitoring and updating,
    \item Building trust with clinicians and hospital staff, 
    \item Adopting interoperability standards, and
    \item Ensuring that deployed models align with ethical considerations, do not exacerbate biases, and adhere to privacy and governance policies 
\end{itemize}

In this review, we articulate the challenges involved in implementing successful Machine Learning Health Operations (MLHOps) pipelines, specific to clinical use cases. We begin by outlining the foundations of model deployment in general, and provide a comprehensive study of the emerging discipline \cite{symeonidis2022mlops,makinen2021needs}. We then provide a detailed review of the different components of development pipelines specific to healthcare. We discuss data, pipeline engineering, deployment, monitoring and updating models, and ethical considerations pertaining to healthcare use cases. While MLHOps often requires aspects specific to healthcare,  best practices and concepts from other application domains are also relevant.  
This summarizes the primary outcome of our review, which is to provide a set of recommendations for implementing MLHOps pipelines in practice -- i.e., a ``how-to" guide for practitioners. 

\section{Foundations of MLOps}
\subsection{What is MLOps?} Machine learning operations (MLOps) is a combination of tools, techniques, standards, and engineering best practices to standardize ML system development and operations \cite{symeonidis2022mlops}. It is used to streamline and automate the deployment, monitoring, and maintenance of machine learning models, in order to ensure they are robust, reliable, and easily updated or upgraded.

\subsection{MLOps Pipeline}
{Pipelines} are processes of multiple modules that streamline the ML workflow. Once the project is defined, the MLOps pipeline begins with identifying the inputs and outputs relevant to the problem, cleaning, and transforming the data towards useful and efficient representations for machine learning, training, and evaluating model performance, and deploying selected models in production while continuing to monitor their performance. Figure \ref{mlops_pipeline} illustrates a general MLOps pipeline. 
Common types of pipelines include:
\begin{itemize}
    \item \textit{Automated pipelines:} An end-to-end pipeline that is automated towards a single task, e.g., a model training pipeline.
    \item \textit{Orchestrated pipelines:} A pipeline that consists of multiple modules, designed for several automated tasks, and managed and coordinated in a dynamic workflow, e.g., the pipeline managing MLOps.
\end{itemize}

Recently, MLOps has become more well-defined and widely implemented due to the reusability and standardization benefits across various applications \cite{ruf2021demystifying}. As a result, the structure and definitions of different components are becoming quite well-established.

\subsection{MLOps Components}
MLOps pipelines consist of different components and key concepts \cite{kreuzberger2022machine,he2021automl}, stated below (and shown in Figure \ref{mlops_pipeline}):

\begin{figure}[ht]
\centering
\includegraphics[width=1.1\textwidth]{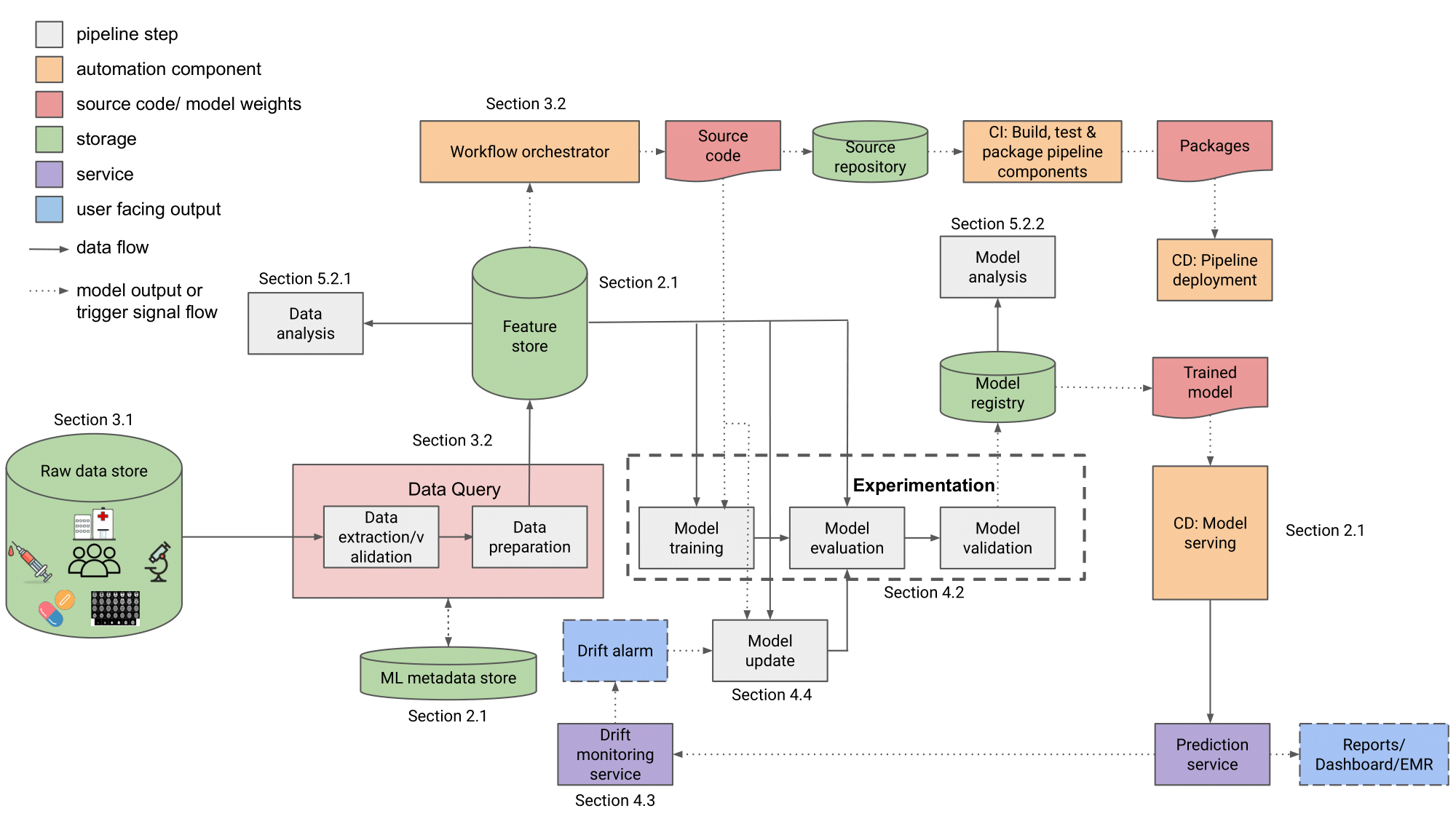}
\caption{MLOps pipeline}
\label{mlops_pipeline}
\end{figure}

\begin{itemize}
    \item \textbf{Stores:} Stores encapsulate the tools designed to centralize building, managing, and sharing either features or models across different teams and applications in an organization. 
\begin{itemize}
    \item \textbf{Raw data source:} A raw data store is a centralized repository that stores data in its raw, unprocessed form. It is a staging area where data is initially collected and stored before processing or transformation.
    \item \textbf{Feature store:} A centralized online repository for storing, managing, and sharing features used in ML models. These features are acquired by processing the raw data and are made available for real-time serving through the feature store.
    
    \item \textbf{ML metadata store:} A ML metadata store helps record and retrieve metadata associated with an ML pipeline including information about various pipeline components, their executions (e.g. training runs), and resulting artifacts (e.g. trained models).
\end{itemize}
    \item \textbf{Serving:} Serving is the task of hosting ML artifacts (usually models) either on the cloud or on-premise so that their functions are accessible to multiple applications through remote function calls (i.e., application programming interfaces (APIs)).

\begin{itemize}
    \item In \textit{batch serving}, the artifact is used by scheduled jobs.
    \item In \textit{online serving}, the artifact processes requests in real-time. Communication and access point channels, traffic management, pre- and post-processing requests, and performance monitoring should all be considered while serving artifacts. 
 
\end{itemize}
    \item \textbf{Data query:} The component queries the data, processes it and stores it in a format that models can easily utilize.  
    \item \textbf{Experimentation:} The experimentation component consists of model training, model evaluation, and model validation.
    \item \textbf{Model registry:} The model registry is a centralized repository that stores trained machine learning models, their metadata, and their versions. 
    \item  \textbf{Drift-detection:} The drift-detection component is responsible for monitoring the AI system for potentially harmful drift and issuing an alert when drift is detected.
    \item \textbf{Workflow orchestration:} The workflow orchestration component responsible for the process of automating and managing the end-to-end flow of the ML pipeline.
    \item \textbf{Source repository:}  The source repository is a centralized code repository that stores the source code (and its history) for ML models and  related components.
    \item \textbf{Containerization:} Containerization involves packaging models with the components required to run them; this includes libraries and frameworks so they can run in isolated user spaces with minimal configuration of the underlying operating system \cite{garg2021continuous}. Sometimes, source code is also included in these containers.
    \end{itemize}

\subsection{Levels of MLOps maturity}
MLOps practices can be divided into different levels based on the maturity of the ML system automation process \cite{9582569,symeonidis2022mlops}, as described below. 

\begin{itemize}
\item \textbf{Level 0 -- Manual ML pipeline:} Every step in the ML pipeline, including data processing, model building, evaluation, and deployment, are manual processes. In Level 0, the experimental and operational pipelines are distinct and the data scientists provide a trained model as an artifact to the engineering team to deploy on their infrastructure. Here, only the trained model is served for deployment and there are infrequent model updates. Level 0 processes typically lack rigorous and continuous performance monitoring capabilities. \newline

\item \textbf{Level 1 -- Continuous Model Training and Delivery:} Here, the entire ML pipeline is automated to perform continuous training of the model as well as continuous delivery of model prediction services. Software orchestrates the execution and transition between the steps in the pipeline, leading to rapid iteration over experiments and an automatic process for deploying a selected model into production. Contrary to Level 0, the entire training pipeline is automated, and the deployed model can incorporate newer data based on pipeline triggers. Given the automated nature of Level 1, it is necessary to continuously monitor, evaluate, and validate models and data to ensure expected performance during production. \newline

\item \textbf{Level 2 -- Continuous Integration and Continuous Delivery:} This involves the highest maturity in automation through enforcing combined practice of continuous integration and delivery which enables for a rapid and reliable update of the pipelines in production. Through automated test and deployment of new pipeline implementations, any rapid changes in data and business environment can be addressed. In this level, the pipeline and its components are automatically built, tested, and packaged when new code is committed or pushed to the source code repository. Moreover, the system continuously delivers new pipeline implementations to the target environment that in turn delivers prediction services of the newly trained model. 
\end{itemize}

Ultimately implementation of MLOps leads to many benefits, including better system quality, increased scalability, simplified management processes, improved governance and compliance, increased cost savings and improved collaboration.

\section{MLHOps Setup}
Operationalizing ML models in healthcare is unique among other application domains. Decisions made in clinical environments have a direct impact on patient outcomes and, hence, the consequences of integrating ML models into health systems need to be carefully controlled. For example, early warning systems might enable clinicians to prescribe treatment plans with increased lead time \cite{henry2022human}; however, these systems might also suffer from a high false alarm rate, which could result in alarm fatigue and possibly worse outcomes. The requirements placed on such ML systems are therefore very high and, if they are not adequately satisfied, the result is diminished adoption and trust from clinical staff. Rigorous long-term evaluation is needed to validate the efficacy and to identify and assess risks, and this evaluation needs to be reported comprehensively and transparently \cite{vasey2022deployment}.\\

While most MLOps best practices extend to healthcare settings, the data, competencies, tools, and model evaluation differ significantly \cite{meng2022interpretability,markus2021role,tonekaboni2022validate,antoniou2021health}. For example, typical performance metrics (e.g., positive predictive value and F1-scores) may differ between clinicians and engineers. Therefore, unlike in other industries, it becomes necessary to evaluate physician experience when predictions and model performance are presented to clinical staff \cite{wang2022physician}. 
In order to build trust in the clinical setting, the interpretability of ML models is also exceptionally important. As more ML models are integrated into hospitals, new legal frameworks and standards for evaluation need to be adopted, and MLHOps tools need to comply with existing standards.


In the following sections, we explore the different components of MLHOps pipelines.

\subsection{Data}
\label{data-handling}
Successfully digitizing health data has resulted in a prodigious increase in the volume and complexity of patient data collected \cite{rajkomar2018scalable}. These datasets are now stored, maintained, and processed by hospital IT infrastructure systems which in turn use specialized software systems. 

\subsubsection{Data sources}

There could be multiple sources of data, which are categorized as follows:

Electronic health records (EHRs) record, analyze, and present information to clinicians, including: 

\begin{enumerate}
    \item \textbf{Patient demographic data:} E.g., age and sex.
    \item \textbf{Administrative data:} E.g., treatment costs and insurance.
    \item \textbf{Patient observations records:} E.g., chart events such as lab tests and vitals. These include a multitude of physiological signals captured using various methods such as heart rate, blood pressure, skin temperature, and respiratory rate.
    \item\textbf{Interventions:} These are steps that significantly alter the course of patient care, such as  mechanical ventilation, dialysis, or blood transfusions.
    \item \textbf{Medications information:} E.g., medications administered and their dosage.
    \item \textbf{Waveform data:} This digitizes physiological signals collected from bedside patient monitors.
    \item \textbf{Imaging reports and metadata:} E.g., CT scans, MRI, ultrasound, and corresponding radiology reports.
    \item \textbf{Medical notes:} These are made by clinical staff on patient condition. These can also be transcribed text of recorded interactions between the patient and clinician.
\end{enumerate}

Other sources of health data include primary care data, wearable data (e.g., smartwatches), genomics data, video data, surveys, medical claims, billing data, registry data, and other patient-generated data \cite{raghupathi2014big,belle2015big,chaudhary2022taxonomy}. \\

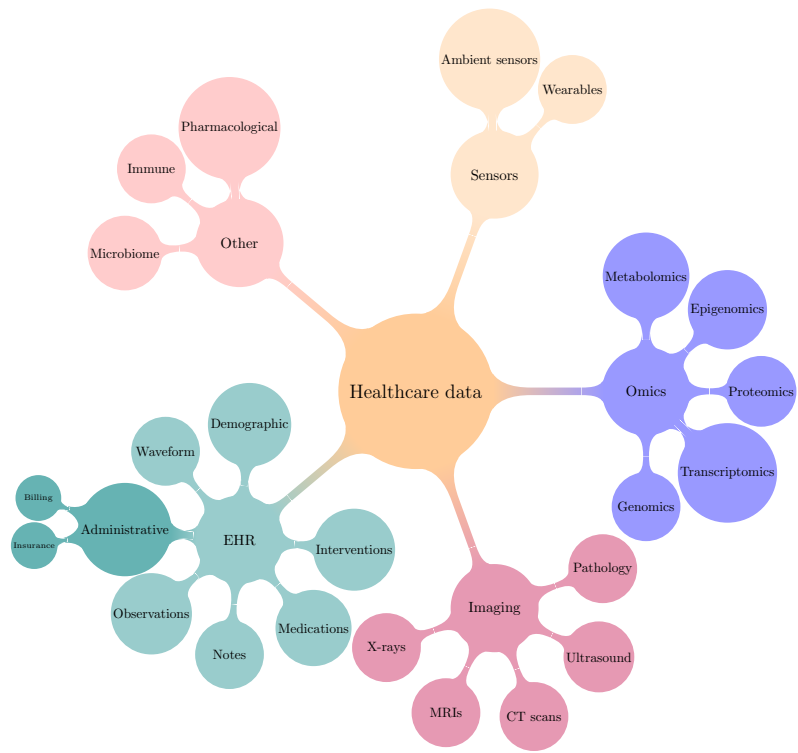
\begin{figure}[ht]
\centering
\resizebox{300px}{280px}{
\begin{tikzpicture}[mindmap, grow cyclic, every node/.style={concept, text width=}, concept color=orange!40, 
	level 1/.append style={level distance=6cm,sibling angle=70},
	level 2/.append style={level distance=3cm,sibling angle=45},
	]
\node{Healthcare data}
child [concept color=teal!40] { node {EHR}
	child { node {Demographic}}
	child { node {Waveform}}
	child [concept color=teal!60] { node {Administrative}
	child { node {Billing}}
	child { node {Insurance}}
	}
	child { node {Observations}}
	child { node {Notes}}
	child { node {Medications}}
	child { node {Interventions}}
}
child [concept color=purple!40] { node {Imaging}
	child { node {X-rays}}
	child { node {MRIs}}
	child { node {CT scans}}
	child { node {Ultrasound}}
	child { node {Pathology}} 
}
child [concept color=blue!40] { node {Omics} 
	child { node {Genomics}}
	child { node {Transcriptomics}}
	child { node {Proteomics}}
	child { node {Epigenomics}}
	child { node {Metabolomics}}
}
child [concept color=orange!20] { node {Sensors} 
	child { node {Wearables}}
	child { node {Ambient sensors}}
}
child [concept color=red!20] { node {Other}
	child { node {Pharmacological}}
	child { node {Immune}}
	child { node {Microbiome}}
};
\end{tikzpicture}
}
\caption{Stratification of health data. Further levels of stratification can be extended as the data becomes richer. For example, observational EHR data could include labs, vital measurements, and other recorded observations.} \label{health_data}
\end{figure}
 
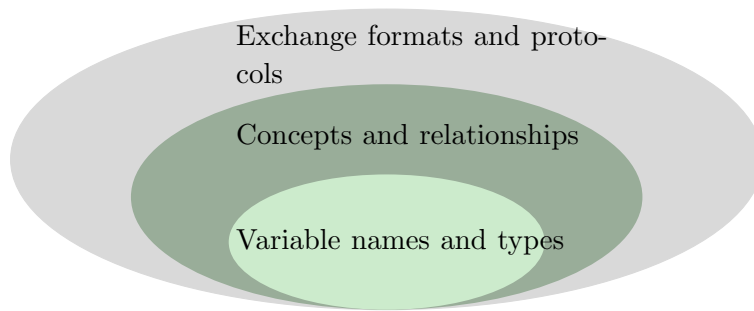
\begin{figure}[ht]
\centering
    \begin{tikzpicture}
    \node[ellipse,
    draw = gray!30,
    text = black,
    minimum width = 10cm, 
    minimum height = 4cm,
    fill = gray!30] (e) at (0, 0) {4.4 and 4.0};
    \fill[fill=black!80!green!40] (0, -0.5) ellipse (3.4 and 1.5);
    \fill[fill=black!40!green!20] (0, -1.1) ellipse (2.1 and 0.9);
    \node[text width=5cm] at (0.5,-1.1) {\small Variable names and types};
    \node[text width=5cm] at (0.5, 0.3) {\small Concepts and relationships};
    \node[text width=5cm] at (0.5, 1.4) {\small Exchange formats and protocols};
    \end{tikzpicture}
\caption{The hierarchy of standardization that common data models and open standards for interoperability address. The lowest level is about achieving standardization of variable names such as lab test names, medications and diagnosis codes, as well as the data types used to store these variables (i.e. integer vs. character). The next level is about having abstract concepts such that data can be mapped and grouped under these concept definitions. The top level of standardization is about data exchange formats, e.g. JSON, XML, protocols, along with protocols for information exchange like supported RESTful API architectures. This level addresses questions on interoperability and how data can be exchanged across sites and EHR systems.} \label{standardization}
\end{figure}

Figure \ref{health_data} illustrates the heterogeneous nature of health data. The stratification shown can be extended further to contain more specialized data. For example, genomics data can be further stratified into different types of data based on the method of sequencing; observational EHR data can be further stratified to include labs, vital measurements, and other recorded observations. \\
\\
With such large volumes and variability in data, standardization is key to achieve scalability and interoperability. Figure \ref{standardization} illustrates the different levels of standardization that need to be achieved with respect to health data.

\subsubsection{Common Data Model (CDM)}

Despite the widespread adoption of EHR systems, clinical events are not captured in a standard format across observational databases \cite{ohdsi2019book}. For effective research and implementation, data must be drawn from many sources and compared and contrasted to be fully understood. \\
 
Databases must also support scaling to large numbers of records which can be processed concurrently. Hence, efficient storage systems along with computational techniques are needed to facilitate analyses. One of the first steps towards scalability is to transform the data to a common data standard. Once available in a common format, the process of extracting, transforming, and loading (ETL) becomes simplified. In addition to scale, patient data require a high level of protection with strict data user agreements and access control. A common data model addresses these challenges by allowing for downstream functional access points to be designed independent of the data model. Data that is available in a common format promotes collaboration and mitigates duplicated effort. Specific implementations and formats of data should be hidden from users, and only high-level abstractions need to be visible. \\

The Systematized Nomenclature of Medicine (SNOMED) was among the first efforts to standardize clinical terminology, and a corresponding dictionary with a broad range of clinical terminology is available as part of SNOMED-CT \cite{donnelly2006snomed}. Several data models use SNOMED-CT as part of their core vocabulary.
Converting datasets to a common data model like the Observational Medical Outcomes Partnership (OMOP) model involves mapping from a source database to the target content delivery manager. This process is usually time-consuming and involves a lot of manual effort undertaken by data scientists. Tools to simplify the mapping and conversion process can save time and effort and promote adoption. For OMOP, the ATLAS tool \cite{ohdsi2019book} developed by Observational Health Data Sciences and Informatics (OHDSI) provides such a feature through their web based interactive analysis platform. 

\subsubsection{Interoperability and open standards}

As the volume of data grows in healthcare institutions and applications ingest data for different use cases, real-time performance and data management is crucial. To enable real-time operation and easy exchange of health data across systems, an interoperability standard for data exchange along with protocols for accessing data through easy-to-use programming interfaces is necessary. Some of the popular healthcare data standards include Health Level 7 (HL7), Fast Healthcare Interoperability Resources (FHIR), Health Level 7 v2 (HL7v2), and Digital Imaging and Communications in Medicine (DICOM). \\

The FHIR standard \cite{fhir} is a leading open standard for exchanging health data. FHIR is developed by Health Level 7 (HL7), a not-for-profit standards development organization that was established to develop standards for hospital information systems. FHIR defines the key entities involved in healthcare information exchange as resources, where each resource is a distinct identifiable entity. FHIR also defines APIs which conform to the representational state transfer (REST) architectural style for exchanging resources, allowing for stateless Hypertext Transfer Protocol (HTTP) methods, and exposing directory-structure like URIs to resources. RESTful architectures are light-weight interfaces that allow for faster transmission, which is more suitable for mobile devices. RESTful interfaces also facilitate faster development cycles because of their simple structure. \\ 

 DICOM is the standard for the communication and management of medical imaging information and related metadata. The DICOM standard specifies the format and protocol for exchange of digital information between medical imaging equipment and other systems. Persistent information objects which encode images are exchanged and an instance of such an information object may be exchanged across many systems and many organizational contexts, and over time. DICOM has enabled deep collaboration and standardization across different disciplines such as radiology, cardiology, pathology, ophthalmology, and related disciplines.

 \subsubsection{Quality assurance and validation}
 
 Data collected in retrospective databases for analysis and ML use cases need to be checked for quality and consistency. Data validation is an important step towards ensuring that ML systems developed using the data are highly performant, and do not incorporate biases from the data. Errors in data propagate through the MLOps pipeline and hence specialized data quality assurance tools and checks at various stages of the pipeline are necessary \cite{renggli2021data}.  A standardized data validation framework that includes i) data element pre-processing, ii) checks for completeness, conformance, and plausibility, and iii) a review process by clinicians and other stakeholders should capture generalizable insight across various clinical investigations \cite{sendak2022development}. 
 

\subsection{Pipeline Engineering}
\label{pipeline_eng}
Data stored in raw formats need to be processed to create feature representations for ML models. 
Each transformation is a computation, and a chain of these processing elements, arranged so that the output of each element is the input of the next, constitutes a {\em pipeline} \cite{kreuzberger2022machine} 
and using software tools and workflow practices that enable such pipelines is {\em pipeline engineering}.\\


There are advantages to using such a pipeline approach, including:
\begin{itemize}
    \item \textbf{Modularization}: By breaking the chain of transformations into small steps, modularization is naturally achieved.
    \item \textbf{Testing}: Each transformation step can be tested independently, which facilitates quality assurance and testing.
    \item \textbf{Debugging}: Version controlling the outputs at each step makes it easier to ensure reproducibility, especially when many steps are involved.
    \item \textbf{Parallelism}: If any step in the pipeline is easily parallelizable across multiple compute nodes, the overall processing time can be reduced.
    \item \textbf{Automation}: By breaking a complex task into a series of smaller tasks, the completion of each task can be used to trigger the start of the next task, and this can be automated using continuous integration tools such as Jenkins, Github actions and Gitlab CI.
\end{itemize}


In health data processing, the following steps are crucial: 

\begin{enumerate}
    \item \textbf{Cleaning}: Formatting values, adjusting data types, checking and fixing issues with raw data.
    \item \textbf{Encoding}: Computing word embeddings for clinical text, encoding the text and raw values into embeddings \cite{khattak2019survey,alsentzer2019publicly}. Encoding is a general transformation step that can be used to create vector representations of raw data. For example, transforming images to numeric representations can also be considered to be encoding.
    \item \textbf{Aggregation}: Grouping values into buckets, e.g., for aggregating measurements into fixed time-intervals, or grouping values by patient ID.
    \item \textbf{Normalization}: Normalizing values into standard ranges or using statistics of the data.
    \item \textbf{Imputation}: Handling missing values in the data. For various clinical data, `missingness' can actually provide valuable contextual information about the patient's health and needs to be handled carefully \cite{che2018recurrent}.
\end{enumerate}

Multiple data sources such as EHR data, clinical notes and text, imaging data, and genomics data can be processed independently to create features and they can be combined to be used as inputs to ML models. Hence, composing pipelines of these tasks facilitates component reusability  \cite{jarrett2020clairvoyance}. Furthermore, since the ML development life-cycle constitutes a chain of tasks, the pipelining approach becomes even more desirable. Some of the high level tasks in the MLHOps pipeline include feature creation, feature selection, model training, evaluation, and monitoring. 
Evaluating models across different slices of data, hyper-parameters, and other confounding variables is necessary for building trust. \\
\\
Table \ref{mlhops_tools} lists popular open-source tools and packages specific to health data and ML processing. These tools are at different stages of development and maturity. Some examples of popular tools include \texttt{MIMIC-Extract} \cite{wang2020mimic}, \texttt{Clairvoyance} \cite{jarrett2020clairvoyance} and \texttt{CheXstray} \cite{soin2022chexstray}.

\subsection{Modelling}
\label{modelling}
At this stage, the data has been collected, cleaned, and curated, ready to be fed to the ML model to accomplish the desired task. The modelling phase involves choosing the available models that fit the problem, training \& testing the models, and choosing the model with the best performance \& reliability guarantees. Given the the existence of numerous surveys summarizing machine learning and deep learning algorithms for general healthcare scenarios \cite{esteva2019guide,abdullah2022review}, as well as specific use cases  such as brain tumor detection \cite{arabahmadi2022deep}, COVID-19 prevention\cite{band2022survey}, and clinical text representation \cite{khattak2019survey}, we omit this discussion and let the reader explore the surveys relevant to their prediction problem.  



\subsection{Infrastructure and System}
\label{infra}
Hospitals typically use models developed by their EHR vendor which are deployed through the native EHR vendor configuration. Often, inference is run locally or in a cloud instance, and the model outputs are communicated within the EHR \cite{kashyap2021deployment}. Predominantly, these models are pre-trained and sometimes fine-tuned on the specific site's data. 

A feature store is a ML-specific data system used to centralize storage, processing, and access to frequently used features, making them available for reuse in the development of future machine learning models. Feature stores operationalize and streamline the input, tracking, and governance of the data as part of feature engineering for machine learning \cite{kreuzberger2022machine}.


To ensure reliability, the development, staging, and production environments are separated and have different requirements. The staging and production environments typically consist of independent virtual machines with adequate compute and storage, along with reliable and secure connections to the databases. 

The infrastructure and software systems also have to follow and comply with cybersecurity, medical software design and software testing standards \cite{de2022deployment}.

\subsubsection{Roles and Responsibilities}
\label{roles}
Efficient and successful MLHOps requires a collaborative, interdisciplinary team across a range of expertise and competencies commonly found in data science, ML, software, operations, production engineering, medicine, and privacy capabilities~\cite{kreuzberger2022machine}. Similar to general MLOps practices, data and ML scientists, data, DevOps, and ML engineers, solution and data architects, ML and software fullstack developers, and project managers are needed. In addition, the following role are required, which are distinct to healthcare (for more general MLOps roles see Table \ref{tab:mlops roles}):
\begin{itemize}    
    \item \textbf{Health AI Project Managers:} Responsibilities include panning projects, establishing guidelines, milestone tracking, managing risk, supporting the teams and governing partnerships  with collaborators from other health organizations.
    \item \textbf{Health AI Implementation Coordinator:} Liaison that engages with key stakeholders to facilitate the implmentation of clinical AI systems.
    \item \textbf{Healthcare Operations Manager:} Oversees and coordinates quality management, resource management, process improvement, and patient safety in clinical settings like hospitals.
    \item \textbf{Clinical Researchers \& Scientists:} Domain experts that provide critical domain-specific knowledge relevant to model development and implementation.
    \item \textbf{Patient-Facing Practitioners:} Responsibilities include providing system requirements, pipeline usage feedback, and perspective about the patient experience (e.g. clinicians, nurses). 
    \item \textbf{Ethicists:} Provides support regarding ethical implications of clinical AI systems.
    \item \textbf{Privacy Analysts:} Provides assessments regarding privacy concerns pertaining to the usage of patient data.
    \item \textbf{Legal Analysts:} Works closely with privacy analysts and ethicists to evaluate the legal vulnerabilities of clinical AI systems.
\end{itemize}    

\subsection{Reporting Guidelines}
\label{reporting}
Many clinical AI systems do not meet reporting standards because of a failure to assess for poor quality or unavailable input data, insufficient analysis of performance errors, or a lack of information regarding code or algorithm availability \cite{plana2022deployment}. Systematic reviews of clinical AI systems suggest there is a substantial reporting burden, and additions regarding reliability and fairness can improve reporting \cite{lu2022reporting}. As a result, guidelines informed by challenges in existing AI deployments in health settings have become imperative \cite{collins2021reporting}. Reporting guidelines including CONSORT-AI \cite{liu2020reporting}, DECIDE-AI \cite{vasey2022deployment}, and SPIRIT-AI \cite{rivera2020reporting} were developed by a multidisciplinary group of international experts using the Delphi process to ensure complete and transparent reporting of randomized clinical trials (RCT) that evaluate interventions with an AI model. Broadly these guidelines suggest inclusion of the following criteria \cite{de2022deployment}:

\begin{itemize} 
    \item \textbf{Intended use}: Inclusion of the medical problem and context, current standard practice, intended patient population(s), how the AI system will be integrated into the care pathway, and the intended patient outcomes.
    \item \textbf{Patient and user recruitment}: Well-defined inclusion and exclusion criteria.
    \item \textbf{Data and outcomes}: The use of a representative patient population, data coding and processing, missing- and low-quality data handling, and sample size considerations.
    \item \textbf{Model}: Inclusion of inputs, outputs, training, model selection, parameter tuning, and performance.
    \item \textbf{Implementation}: Inclusion of user experience with the AI system, user adherence to intended implementation, and changes to clinical workflow.
    \item \textbf{Modifications}: A description protocol for changes made, timing and rationale for modifications, and outcome changes after each modification.
    \item \textbf{Safety and errors}: Identification of system errors and malfunctions, anticipated risks and mitigation strategies, undesirable outcomes, and worst-case scenarios.
    \item \textbf{Ethics and fairness}: Inclusion of subgroup analyses, and fairness metrics.  
    \item \textbf{Human-computer agreement}: Report of user agreement with the AI system, reasons for disagreement, and cases of users changing their mind based on the AI system.
    \item \textbf{Transparency}: Inclusion of data and code availability.
    \item \textbf{Reliability}: Inclusion of uncertainty measures, and performance against realistic baselines.
    \item \textbf{Generalizability}: Inclusion of measures taken to reduce overfitting, and external performance evaluations.
\end{itemize}

\subsubsection{Tools and Frameworks}
Understanding the MLOps pipeline and required expertise is just the first step to addressing the problem. Once this has been accomplished, it is necessary to create and/or adopt appropriate tooling for transforming these principles into practice. There are seven broad categories of MLOps tools as shown in Table~\ref{tab:mlops tools} whereby different tools to automate different phase of the workflows involved in MLOps processes exist. A compiled list of tools within each category is shown in Table~\ref{tab:mlops tools}

\begin{table}
\centering
\caption{MLOps tools}
\label{tab:mlops tools}
\begin{tabular}{p{5cm}|p{5cm}|p{4cm}}
  \hlineB{4}
{\textbf{Category}} & \multicolumn{1}{c|}{\textbf{Description}} & \textbf{Tooling Examples}  \\ 
\hlineB{4}
Model metadata storage and management  & \textbf{Section} \ref{data-handling}  & 

\begin{tabular}{@{\labelitemi\hspace{\dimexpr\labelsep+0.5\tabcolsep}}l@{}}\textit{MLFlow \footnote{\url{https://mlflow.org/docs/latest/index.html}}}\\\textit{Comet \footnote{\url{https://mlops.community/learn/metadata-storage-and-management/comet-ml/}}}\\\textit{Neptune\footnote{\url{https://docs.aws.amazon.com/neptune/latest/userguide/feature-overview-storage.html}}}\\\end{tabular} \\ 
\cline{1-2}
Data and pipeline versioning           &  \textbf{Section} \ref{pipeline_eng}

& 
\begin{tabular}{@{\labelitemi\hspace{\dimexpr\labelsep+0.5\tabcolsep}}l@{}}\textit{DVC\footnote{\url{https://dvc.org/doc/start/data-management/data-pipelines}}}\\\textit{Pachyderm\footnote{\url{https://www.jetbrains.com/pycharm/}}}\\\end{tabular} \\
\cline{1-2}
Model deployment and serving           &  \textbf{Section} \ref{modelling} &  

\begin{tabular}{@{\labelitemi\hspace{\dimexpr\labelsep+0.5\tabcolsep}}l@{}}\textit{DEPLOYR\footnote{\url{https://www.deployr.ai/home-en/}}}\cite{corbin2023deployr}\\\textit{Flyte\footnote{\url{https://flyte.org/}}}\\\textit{ZenML\footnote{\url{https://zenml.io/home}}}\\\end{tabular} \\
\cline{1-2}
Production model monitoring           &  \textbf{Section} \ref{monitor-update} &    \begin{tabular}{@{\labelitemi\hspace{\dimexpr\labelsep+0.5\tabcolsep}}l@{}}\textit{MetaFlow\footnote{\url{https://metaflow.org/}}}\\\textit{Kedro\footnote{\url{https://docs.kedro.org/en/stable/index.html}}}\\\textit{Seldon Core\footnote{\url{https://www.seldon.io/solutions/open-source-projects/core}}}\\\end{tabular} \\
\cline{1-2}
Run orchestration and workflow pipelines           &  Orchestrating the execution of preprocessing, training, and evaluation pipelines. \textbf{Section} \ref{infra} \& \ref{reporting} &   \begin{tabular}{@{\labelitemi\hspace{\dimexpr\labelsep+0.5\tabcolsep}}l@{}}\textit{Kuberflow\footnote{\url{https://www.kubeflow.org/}}}\\\textit{Polyaxon\footnote{\url{https://polyaxon.com/}}}\\\textit{MLRun\footnote{\url{https://www.mlrun.org/}}}\\\end{tabular}                     \\
\hline

Collaboration Tool  &   Setting up an MLOps pipeline requires collabora-
tion between different people. \textbf{Section} \ref{roles} &

\begin{tabular}{@{\labelitemi\hspace{\dimexpr\labelsep+0.5\tabcolsep}}l@{}}\textit{ChatOps\footnote{\url{https://docs.gitlab.com/ee/ci/chatops/}}}\\\textit{Slack\footnote{\url{https://slack.com/}}}\\\textit{Trello\footnote{\url{https://trello.com/}}}\\

\textit{GitLab\footnote{\url{https://about.gitlab.com/}}}\\
\textit{Rocket Chat  \footnote{\url{https://rocket.chat/}}}\end{tabular} \\
\hline
\end{tabular}
\end{table}

\section{MLHOps Monitoring and Updating} 

\label{monitor-update}
Once an MLHOps pipeline and required resources are setup and deployed, robust monitoring protocols are crucial to the safety and longevity of clinical AI systems. For example, inevitable updates to a model can introduce various operational issues (and vice versa), including bias (e.g., a new hospital policy that shifts the nature of new data) and new classes (e.g., new subtypes in a disease classifier) \cite{yoshida2018monitor}. Incorporating expert labels can improve model performance; however, the time, cost, and expertise required to acquire accurate labels for very large imaging datasets like those used in radiology- or histology-based classifiers makes this difficult \cite{lee2020deployment}. \\

As a result, there exist monitoring frameworks with policies to determine when to query experts for labels  \cite{zhao2021mlops}. These include:
\begin{itemize}
    \item \textbf{Periodic Querying}, a non-adaptive policy whereby labels are periodically queried in batches according to a predetermined schedule;
    \item \textbf{Request-and-Reverify} which  sets a predetermined threshold for drift and queries a batch of labels whenever the drift threshold is exceeded \cite{yu2018monitoring};
    \item \textbf{MLDemon} which follows a periodic query cycle and uses a linear estimate of the accuracy based on changes in the data \cite{ginart2022monitoring}. 
\end{itemize}
  
\subsection{Time-scale windows} Monitoring clinical AI systems requires evaluating robustness to temporal shifts. Since the time-scale used can change the types of shifts detected (i.e., gradual versus sudden shifts), multiple time windows should be considered (e.g., week, month). Moreover, it is important to use both 1) \textbf{cumulative  statistics}, which use a single time window and updates at the beginning of each window and 2) \textbf{sliding statistics}, which retain previous data and update with new data.

\subsection{Appropriate metrics} It is critical to choose evaluation and monitoring metrics optimal for each clinical context. The quality of labels is highly dependent on the data from which they are derived and, as such, can possess inherent biases. For instance, sepsis labels derived from incorrect billing codes will inherently have a low positive predictive value (PPV). Moreover, clinical datasets are often imbalanced, consisting of far fewer positive instances of a label than negative ones. As a result, measures like accuracy that weigh positive and negative labels equally can be detrimental to monitoring. For instance, in the context of disease classification, it may be particularly important to monitor sensitivity, in contrast to more time-sensitive clinical scenarios like the intensive care unit (ICU) where false positives (FP) can have critical outcomes \cite{bedsbench2021drift}.

\subsection{Detecting data distribution shift}
Data distribution shift occurs when the underlying distribution of the training data used to build an ML model differs from the distribution of data applied to the model during deployment \cite{candela2009shift}. When the difference between the probability distributions of these data sets is sufficient to deteriorate the model’s performance, the shift is considered malignant. 

In healthcare, there are multiple sources of data distribution shifts, many of which can occur concurrently \cite{finlayson2021shift, subasri2023diagnosing}. Common occurrences of malignant shifts include differences attributed to:
\begin{itemize}
    \item \textbf{Institution} -  These differences can arise when comparing teaching to non-teaching hospitals, government-owned to private hospitals, or general to specialized hospitals (e.g., paediatric, rehabilitation, trauma). These institutions can have differing local clinical practices, resource allocation schemes, medical instruments, and data-collection and processing workflows that can lead to downstream variation. This has previously been reported in Pneumothorax classifiers when evaluated on external institutions \cite{kitamura2020retraining}.
    \item \textbf{Behaviour} - Temporal changes in behaviour at the systemic, physician and patient levels are unavoidable sources of data drift. These changes include new healthcare reimbursement incentives, changes in the standard-of-care in medical practice, novel therapies, and updates to hospital operational processes. An example of this is the COVID-19 pandemic, which required changes in resource allocation to cope with hospital bed shortages \cite{knaus2022shift, park2021drift}. 
    \item \textbf{Patient demographics} - Differences in factors like age, race, gender, religion, and socioeconomic background can arise for various reasons including epidemiological transitions, gentrification of neighbourhoods around a health system, and new public health and immigration policies. Distribution shifts due to demographic differences can disproportionately deteriorate model performance in specific patient populations. For instance, although Black women are more likely to develop breast tumours with poor prognosis, many breast mammography ML classifiers experience deterioration in performance on this patient population \cite{yala2019fairness}. Similarly, skin-lesion classifiers trained primarily on images of lighter skin tones may show decreased performance when evaluated on images of darker skin tones \cite{adamson2018shift,duharpur2021shift}.
    \item \textbf{Technology} - Data shifts can be attributed to changes in technology between institutions or over time. This includes chest X-ray classifiers trained on portable radiographs that are evaluated on stationary radiographs or deteroriation of clinical AI systems across EHR systems (e.g., Philips Carevue vs. Metavision) \cite{nestor2019deployment}.
\end{itemize}

Although evaluated differently, data shifts are present across various modalities of clinical data such as medical images \cite{xiaoyuan2021drift} and EHR data \cite{duckworth2021drift,park2021drift}. In order to effectively prevent these malignant shifts from occurring, it is necessary to perform prospective evaluation of clinical AI systems \cite{otles2021drift} in order to understand the circumstances under which they arise, and to design strategies that mitigate model biases and improve models for future iterations \cite{zhang2022deployment}. Broadly, these data shifts can be categorized into three groups which can co-occur or lead to one another:

\subsubsection{Covariate Shift} Covariate shift is a difference in the distribution of input variables between source and target data. It can occur due to a lack of randomness, inadequate sampling, biased sampling, or a non-stationary environment. This can be at the level of a single input variable (i.e. feature shift) or a group of input features (i.e. dataset shift). Table \ref{covariate_shift_detection_methods} contains a list of commonly used methods used for covariate shift detection.\\
\\    
\textbf{Feature Shift Detection:} Feature shift refers to the change in distribution between the source and target data for a single input feature. Feature shift detection can be performed using two-sample univariate tests such as the Kolmogorov-Smirnov (KS) test \cite{rabanser2019drift}. Publicly available tools like TensorFlow Extended (TFX) apply univariate tests (i.e., $L$-infinity distance for categorical variables, Jensen-Shannon divergence for continuous variables) to perform feature shift detection between training and deployment data and provide users with summary statistics (Table \ref{monitor_update_tools}). It is also possible to detect feature shift while conditioning on the other features in a model using conditional distribution tests \cite{kulinski2020drift}. \\
\\
\textbf{Dataset Shift Detection:} Dataset shift refers to the change in the joint distribution between the source and target data for a group of input features. Multivariate testing is crucial because input to ML models typically consist of more than one variable and multiple modalities. In order to test whether the distribution of the target data has drifted from the source data two main approaches exist: 1) \textbf{two-sample testing} and 2) \textbf{classifiers}. These approaches often work better on low-dimensional data compared to high-dimensional data, therefore dimensionality reduction is typically applied first \cite{rabanser2019drift}. For instance, variational autoencoders (VAE) have been used to reduce chest X-ray images to a low-dimensional space prior to two-sample testing \cite{soin2022chexstray}. In the context of medical images, including  chest X-rays \cite{pooch2019shift} \cite{zech2018shift}, diabetic retinopathies \cite{cao2020drift}, and histology slides \cite{stacke2020shift}, classifier methods have proven effective. For EHR data, dimensionality reduction using clinically meaningful patient representations has improved model performance \cite{nestor2019deployment}. For clinically relevant drift detection, it is important to ensure that drift metrics correlate well with ground truth performance differences. \\

\begin{table}[ht]
\begin{center}
\begin{tabular}{ |p{7.8cm}||p{2.2cm}||p{1.8cm}| }
\hline
Method & Shift & Test Type \\
\hline
L-infinity distance & Feature (c) & 2-ST \\
Cramér-von Mises & Feature (c) & 2-ST\\
Fisher's Exact Test & Feature (c) & 2-ST \\
Chi-Squared Test & Feature (c) & 2-ST \\
Jensen-Shannon divergence & Feature (n) & 2-ST \\
Kolmogorov-Smirnov \cite{massey1951kolmogorov} & Feature (n) & 2-ST \\
Feature Shift Detector \cite{kulinski2020drift} & Feature & Model  \\
Maximum Mean Discrepancy (MMD) \cite{gretton2012shift} & Dataset & 2-ST \\
Least Squares Density Difference \cite{bu2016shift}& Dataset & 2-ST \\ 
Learned Kernel MMD \cite{liu2021shift} & Dataset & 2-ST \\
Context Aware MMD \cite{cobb2022shift} & Dataset & 2-ST \\
MMD Aggregated \cite{schrab2021shift} & Dataset & 2-ST \\ 
Classifier \cite{lopez2016shift} & Dataset & Model \\
Spot-the-diff \cite{jitkrittum2016shift} & Dataset & Model \\
Model Uncertainty \cite{sethi2017shift} & Dataset & Model \\
Mahalanobis distance \cite{ren2021shift} & Dataset & Model \\
Gram matrices \cite{park2020drift} \cite{sastry2020shift} & Dataset &  Model \\
Energy Based Test \cite{liu2020energy}& Dataset &  Model \\
H-Divergence \cite{zhao2021shift}& Dataset & Model \\
\hline
\end{tabular}
\end{center}
\label{covariate_shift_detection_methods}
\caption{\textbf{Covariate Shift Detection Methods} c: categorical; n: numeric; 2-ST: Two-Sample Test}
\end{table}

\subsubsection{Concept Shift} Concept shift is a difference in the relationship (i.e., joint distribution) of the variables and the outcome between the source and target data. In healthcare, concept shift can arise due to changes in symptoms for a disease or antigenic drift. This has been explored in the context of surgery prediction \cite{beyene2015drift} and medical triage for emergency and urgent care \cite{huggard2020drift}.\\

\textbf{Concept Shift Detection:} There are three broad categories of concept shift detection based on their approach.
\begin{enumerate}
    \item \textbf{Distribution techniques} which use a sliding window to divide the incoming data streams into windows based on data size or time interval and that compare the performance of the most recent observations with a reference window  \cite{gama2014survey}. ADaptive WINdowing (ADWIN), and its extension ADWIN2, are windowing techniques which use the Hoeffding bound to examine the change between the means of two sufficiently large subwindows \cite{moharram2022drift}.
    \item \textbf{Sequential Analysis} strategies use the Sequential Probability Ratio Test (SPRT) as the basis for their change detection algorithms. A well-known algorithm is CUMSUM which outputs an alarm when the mean of the incoming data significantly deviates from zero \cite{bayram2022concept}.
    \item \textbf{Statistical Process Control} (SPC) methods track changes in the online error rate of classifiers and trigger an update process when there is a statistically significant change in error rate \cite{lu2020drift}. Some common SPC methods include: Drift Detection Method (DDM), Early Drift Detection Method (EDDM), and Local Drift Detection (LLDD) \cite{baenagarcia2006drift}.
\end{enumerate}

\subsubsection{Label Shift} Label shift is a difference in the distribution of class variables in the outcome between the source and target data. Label shift may appear when some concepts are under-sampled or over-sampled in the target domain compared to the source domain. Label shift arises when class proportions differ between the source and target, but the feature distributions of each class do not. For instance, in the context of disease diagnosis, a classifier trained to predict disease occurrence is subject to drift due to changes in the baseline prevalence of the disease across various populations. \\
\\
\textbf{Label Shift Detection:} Label shift can be detected using moment matching-based estimator methods that leverage model predictions like Black Box Shift Estimation (BBSE) \cite{lipton2018drift} and Regularized Learning under Label Shift (RLLS) \cite{azizzadenesheli2019drift}. Assuming access to a classifier that outputs the true source distribution conditional probabilities \begin{math} p_s(y|x) \end{math} Expectation Maximization (EM) algorithms like Maximum Likelihood Label Shift (MLLS) can also be used to detect label shift \cite{garg2020drift}. Furthermore, methods using bias-corrected calibration show promise in correcting label shift \cite{alexandari2020shift}.

\subsection{Model Updating and Retraining}

As the implementation of ML-enabled tools is realized in the clinic, there is a growing need for continuous monitoring and updating in order to improve models over time and adapt to malignant distribution shifts. Retraining of ML models has demonstrated improved model performance in clinical contexts like pneumothorax diagnosis \cite{kitamura2020retraining}. However, proposed modifications can also degrade performance and introduce bias \cite{liley2021update}; as a result it may be preferable to avoid making a prediction and defer the decision to a downstream expert \cite{mozannar2020monitor}. When defining a model updating or retraining strategy for clinical AI models there are several factors to consider \cite{deltagrad2020retraining}, we outline they key criteria in this section. 

\begin{table}[ht]
\begin{center}
\begin{tabular}{ |p{4.5cm}||p{8cm}| }
 \hline
 Name of tool &  Capabilities \\
 \hline
 \texttt{Evidently} \footnote{\url{https://github.com/evidentlyai/evidently}} &  Interactive reports to analyze ML models during validation or production monitoring. \\
 \texttt{NannyML}\footnote{\url{https://github.com/NannyML/nannyml}} &  Performance estimation and monitoring, data drift detection and intelligent alerting for deployment.\\
 \texttt{River} \cite{montiel2021river} & Online metrics, drift detection and outlier detection for streaming data. \\
 \texttt{SeldonCore} \cite{van2019alibi} & Serving, monitoring, explaining, and management of models using advanced metrics, explainers, and outlier detection. \\
 \texttt{TFX}\footnote{\url{https://github.com/tensorflow/tfx}} &  Explore and validate data used for machine learning models. \\
\texttt{TorchDrift}\footnote{\url{https://github.com/TorchDrift/TorchDrift}} & Covariate and concept drift detection. \\
 \texttt{deepchecks} \cite{deepchecks} & Testing for continuous validation of ML models and data. \\
 \texttt{EHR OOD Detection} \cite{ulmer2020trust} & Uncertainty estimation, OOD detection and (deep) generative modelling for EHRs. \\
 \texttt{Avalanche} \cite{lomonaco2021avalanche} &  Prototyping, training and reproducible evaluation of continual learning algorithms. \\
 \texttt{Giskard}\footnote{\url{https://github.com/Giskard-AI/giskard}} & Evaluation, monitoring and drift testing. \\
 \hline
\end{tabular}
\end{center}
\label{monitor_update_tools}
\caption{List of open-source tools available on Github that can be used for ML Monitoring and Updating}
\end{table}

\subsubsection{Quality and Selection of Model Update Data} When updating a model it is important to consider the relevance and size of the data to be used. This is typically done by defining a window of data to update the model: i) \textbf{Fixed window} uses a window that remains constant across time. ii) \textbf{Dynamic window} uses a window that changes in size due to an adaptive data shift, iii) \textbf{Representative subsample} uses a subsample from a window that is representative of the entire window distribution.

\subsubsection{Updating Strategies} There are several ways to update a model including: i) \textbf{Model recalibration} is the simplest type of model update, where continuous scores (e.g. predicted risks) produced by the original model are mapped to new values \cite{chen2018calibration}. Some common methods to achieve this include Platt scaling \cite{platt1999probabilistic}, temperature scaling, and isotonic regression  \cite{niculescu2005predicting}. ii) \textbf{Model updating} includes changes to an existing model, for instance, fine-tuning with regularization \cite{lee2019mixout} or model editing where pre-collected errors are used to train hypernetworks that can be used to edit a model’s behaviour by predicting new weights or building a new classifier \cite{mitchell2022memory}. iii) \textbf{Model retraining} involves retraining a model from scratch or fitting an entirely different model. 

\subsubsection{Frequency of Model Updates} In practice, retraining procedures for clinical AI models have generally been locked after FDA approval \cite{lee2021continual} or confined to ad-hoc one-time updates \cite{van2014recalibration} \cite{harrison2006recalibration}. The timing of when it is necessary to update or retrain a model varies across use case. As a result, it is imperative to evaluate the appropriate frequency to update a model. Strategies employed include: i) \textbf{Periodic training} on a regular schedule (e.g. weekly, monthly). ii) \textbf{Performance-based trigger} in response to a statistically significant change in performance. iii) \textbf{Data-based trigger} in response to a statistically significant data distribution shift. iv) \textbf{Retraining on demand} is not based on a trigger or regular schedule, and instead initiated based on user prompts. 

\subsubsection{Continual Learning} Continual learning is a strategy used to update models when there is a continuous stream of input data that may be subject to changes over time. Prior to deployment, it is crucial to simulate the online learning procedure on retrospective data to assess robustness to data shifts \cite{chen2017decaying} \cite{paleyes2022deployment}. When models are retrained on only the most recent data, this can result in “catastrophic forgetting” \cite{vokinger2021continual} \cite{lee2021continual}, in which the integration of new data into the model can overwrite knowledge learned in the past and interfere with what the model has already learned \cite{lee2020deployment}. Contrastingly, procedures that retrain models on all previously collected data can fail to adapt to important temporal shifts and are computationally expensive. More recently, strategies leveraging \textbf{multi-armed bandits} have been utilized to select important samples or batches of data for retraining \cite{graves2017mab} \cite{zhu2021continual}. This is an important consideration in healthcare contexts like radiology, where the labelling of new data can be a time-consuming bottleneck \cite{hadsell2020retraining} \cite{pianykh2020continuous}. \\ 
\\To ensure continual learning satisfies performance guarantees, hypothesis testing can be used for approving proposed modifications \cite{davis2019nonparametric}. An effective approach for parametric models include continual updating procedures like online recalibration/revision \cite{feng2022clinical}. Strategies for continual learning can broadly be categorized into: 1) \textbf{Parameter isolation} where changes to parameters that are important for the previous tasks are forbidden e.g. Local Winner Takes All (LWTA), Incremental Moment Matching (IMM) \cite{van2019learning}; 2) \textbf{Regularization methods} which builds on the observation forgetting can be reduced by protecting parameters that are important for the previous tasks e.g. elastic weight consolidation (EWC), Learning Without Forgetting (LWF); and 3) \textbf{Replay-based approaches} that retain some samples from the previous tasks and use them for training or as constraints to reduce forgetting e.g. episodic representation replay (ERR) \cite{de2021continual}. Evaluation of several continual learning methods on ICU data across a large sequence of tasks indicate replay-based methods achieves more stable long-term performance, compared to regularization and rehearsal based methods \cite{armstrong2022continual}. In the context of chest X-ray classification, Joint Training (JT) has demonstrated superior model performance, with LWF as a promising alternative in the event that training data is unavailable at deployment \cite{lenga2020continual}. For sepsis prediction using EHR data, a joint framework leveraging EWC and ERR  has been proposed \cite{amrollahi2022continual}. More recently, continual model editing strategies have shown promise in overcoming the limitations of continual fine-tuning methods by updating model behavior with minimal influence on unrelated inputs and maintaining upstream test performance \cite{hartvigsen2022editing}. 

\subsubsection{Domain Generalization and Adaptation} Broadly, domain generalization and adaptation methods are used to improve clinical AI model stability and robustness to data shifts by reducing distribution differences between training and test data \cite{zhang2021domain} \cite{guan2021domain}. However, it is critical to evaluate several methods over a range of metrics, as the effectiveness of each method varies based on several factors including the type of shift and data modality \cite{yang2023shift}. 
\begin{itemize}
    \item \textbf{Data-based methods} perform manipulations based on the patient data to minimize distribution shifts. This can be done by re-weighting observations during training based on the target domain \cite{kouw2019domain}, upsampling informative training examples \cite{liu2021retrain} or leveraging a combination of labeled and pseudo-labeled \cite{liao2021domain}. 
    \item \textbf{Representation-based methods} focus on achieving a feature representation such that the source classifier performs well on the target domain. In clinical data this has been explored using strategies including invariant risk minimization (IRM), distribution matching (e.g. CORAL) and domain-adversarial adaptation networks (DANN). DANN methods have demonstrated a reduction on the impact of data shift on cross-institutional transfer performance for diagnostic prediction \cite{zhang2022domain}. However, it has been shown that for clinical AI models subject to real life data shifts, in contrast to synthetic perturbations, empirical risk minimization outperforms domain generalization and unsupervised domain adaptation methods \cite{guo2022drift} \cite{zhang2021empirical}. \item \textbf{Inference-based methods} introduce constraints on the optimization procedure to reduce domain shift \cite{kouw2019domain}. This can be done by estimating a model's performance on the “worst-case” distribution \cite{subbaswamy2021drift} or constraining the learning objective to enforces closeness between protected groups \cite{schrouff2022shift}. Batch normalization statistics can also be leveraged to build models that are more robust to covariate shifts \cite{schneider2020covariate}. 
\end{itemize}
\subsubsection{Data Deletion and Unlearning} In healthcare there are two primary reasons for wanting to remove data from models. Firstly, with the growing concerns around privacy and ML in healthcare, it may become necessary to remove patient data for privacy reasons. Secondly, it may also be beneficial to a model's performance to delete noisy or corrupted training data \cite{bourtoule2021unlearning}. The naive approach to data deletion is to exclude unwanted samples and retrain the model from scratch on the remaining data, however this approach can quickly become time consuming and resource-intensive \cite{izzo2021retraining}. As a result, more sophisticated approaches have been proposed for unlearning in linear and logistic models \cite{izzo2021retraining}, random forest models \cite{brophy2021unlearning}, and other non-linear models \cite{guo2019certified}. 

\subsubsection{Feedback Loops} Feedback loops that incorporate patient outcomes and clinician decisions are critical to improving outcomes in future model iterations. However, retraining feedback loops can also lead to error amplification, and subsequent downstream increases in false positives \cite{adam2020deployment}. As a result, it is important to consider model complexity and choose an appropriate classification threshold to ensure minimization of error amplification \cite{adam2022deployment}. 



\section{Responsible MLHOps}
AI has surged in healthcare, out of necessity or/and \cite{zhang2022deployment,pannu2015ai}, but many issues still exist. For instance, many sources of bias exist in clinical data, large models are opaque, and there are malicious actors who damage or pollute the AI/ML systems. In response, {\em responsible AI} and trustworthiness have together become a growing area of study \cite{mehrabi2021survey,varshney2019trustworthy}. Responsible AI, or trustworthy MLOps, is defined as an ML pipeline that is fair and unbiased, explainable and interpretable, secure, private, reliable, robust, and resilient to attacks. In healthcare, trust is critical to ensuring a meaningful relationship between the healthcare provider and patient \cite{dawson2015trust}.
In this section, we discuss components of responsible and trustworthy AI \cite{li2021trustworthy}, which can be applied to the MLHOps pipeline. In Section \ref{defresAI}, we review the main concepts of responsible AI and in Section \ref{resmlhops} we explore how these concepts can be embedded in the MLHOps pipeline to enable safe deployment of clinical AI systems. \\

\subsection{Responsible AI in healthcare}
\label{defresAI} 
\textbf{Ethics in healthcare:}

Ethics in healthcare primarily consists of the following criteria \cite{varkey2021principles}:

\begin{enumerate}
     \item \textbf{Nonmaleficence:} Do not harm the patient.
    \item \textbf{Beneficence:} Act to the benefit of the patient.
    \item \textbf{Autonomy:} The patient (when able) should have the freedom to make decisions about his/her body. More specifically, the following aspects should be taken care of:
    \begin{itemize}
    \item \textbf{Informed Consent:} The patient (when able) should give informed consent for any medical or surgical procedure, or for research.
    \item \textbf{Truth-telling:} The patient (when able) should receive full disclosure to his/her diagnosis and prognosis. 
    \item \textbf{Confidentiality:} The patient's medical information should not be disclosed to any third party without the patient's consent.
    \end{itemize}
    \item \textbf{Justice:} Ensure fairness to the patient.
\end{enumerate}

To supplement these criteria, guiding principles drawn from surgical settings \cite{little2001invited,rudzicz2020ethics} include:

\begin{enumerate}\addtocounter{enumi}{4}
    \item \textbf{Rescue:}  A patient surrenders to the healthcare provider's expertise to be rescued.
    \item \textbf{Proximity:} The emotional proximity to the patient should be limited to maintain self-preservation and stability in case of any failure. 
    \item \textbf{Ordeal:} A patient may have to face an ordeal (i.e., go through painful procedures) in order to be rescued.
    \item \textbf{Aftermath:} The  physical and psychological aftermath that may occur to the patient due to any treatment must be acknowledged.
    \item \textbf{Presence:} An empathetic presence must be provided to the patient. 
\end{enumerate}

While some of these criteria relate to the humanity of the healthcare provider, others relate to the following topics in ML models:

\begin{itemize}
 \renewcommand{\labelitemi}{\scriptsize$\blacksquare$}

    \item \textbf{Fairness} involves the justice component in the healthcare domain
 \cite{chen2021ethical}.
    \item \textbf{Interpretability \& explainability} relate to explanations and better understanding of the ML models' decisions, which can help in achieving nonmaleficence, beneficence, informed consent, and truth-telling principles in healthcare. Interpretability can help identify the reasons for a given model outcome, which can help inform healthcare providers and patients on how to respond accordingly \cite{meng2022interpretability}.
    \item \textbf{Privacy and security} relate to confidentiality. \cite{kaye2012tension}.

    \item \textbf{Reliability, robustness, and resilience} addresses rescue \cite{rotteau2022striving}.
\end{itemize}
 We discuss these concepts further in Sections \ref{bias_fairness}, \ref{interpretability}, \ref{privacy} and \ref{3r}. 

\subsubsection{Bias \& Fairness}
\label{bias_fairness}

The fairness of AI-based decision support systems have been studied generally in a variety of applications including occupation classifiers \cite{de2019bias},  criminal risk assessments algorithms \cite{chouldechova_fair_2016}, recommendation systems \cite{FairRec}, facial recognition algorithms \cite{shade18}, search engines \cite{FairSerchEng}, and risk score assessment tools in hospitals \cite{racehealth19}. In recent years, the topic of fairness in AI models in healthcare has received a lot of attention \cite{racehealth19, CheXclusion_2020, Larrazabal2020GenderII, Irene2020, wiens2019no, Underdiagnosis_2021}. Unfairness in healthcare manifests as differences in model performance against or in favour of a sub-population, for a given predictive task. For instance, disproportionate performance differences for disease diagnosis in Black versus White patients \cite{CheXclusion_2020}. \\

\paragraph{Causes}
A lack of fairness in clinical AI systems may be a result of various contributing causes:
\begin{itemize}

   \item \textbf{Objective:} 
     \begin{itemize}
  
      \item \textbf{Unfair objective functions:}    
        The initial objective used in developing a ML approach may not consider fairness. 
        This does not mean that the developer explicitly (or implicitly) used an unfair objective function to train the model, but the oversimplification of that objective can lead to downstream issues. For example, a model designed to maximize accuracy across all populations may not inherently provide fairness across different sub-populations even if it reaches state-of-the-art performance on average, across the whole population \cite{CheXclusion_2020, Underdiagnosis_2021}.  
      \item \textbf{Incorrect presumptions:} In some instances, the objective function includes incorrect interpretations of features, which can lead to bias. For instance, a commercial algorithm used in the USA, used health costs as a proxy for health needs\cite{racehealth19}; however, due to  financial limitations, Black patients with the same need for care as White patients often spend less on healthcare and therefore have a lower health cost. As a result, the model falsely inferred that Black patients require less care compared to White patients because they spend less \cite{racehealth19}. Additionally, patients may be charged differently for the same service based on their insurance, suggesting cost may not be representative of healthcare needs.
      \end{itemize}  
        
   \item \textbf{Data:}
       \begin{itemize}

        \item \textbf{Inclusion and exclusion}:
            It is important to clearly outline the conditions and procedures utilized for patient data collection, in order to understand patient inclusion criteria and any potential selection biases that could occur. For instance, the  Chest X-ray dataset \cite{wang2017chestx} was gathered in a research hospital that does not routinely conduct diagnostic and treatment procedures\footnote{from \url{ https://clinicalcenter.nih.gov/about/welcome/faq.html}.}. This dataset therefore includes mostly critical cases, and few patients at the early stages of diagnosis. Moreover, as a specialized hospital, patient admission is selective and chosen solely by institute physicians based on if they have an illness being studied by the given institute \footnote{from \url{https://clinicalcenter.nih.gov/about/welcome/faq.html}.}. Such a dataset will not contain the diversity of disease cases that might be seen in hospitals specialized across different diseases, or account for patients visiting for routine treatment services at general hospitals. 
       
        \item \textbf{Insufficient sample size:}
            Insufficient sample sizes of under-represented groups can also result in unfairness \cite{biasEHR}. For instance, patients of low socioeconomic status may use healthcare services less, which reduces their sample size in the overall dataset, resulting in an unfair model \cite{zhang2021empirical, shade18, Irene2020}.
            In another instance, an algorithm that can classify skin cancer \cite{esteva_dermatologist-level_2017} with high accuracy will not be able to generalize to different skin colours if similar samples have not been represented sufficiently in the training data \cite{shade18}.

        \item \textbf{Missing essential representative features}:
         Sometimes, essential representative features are missed or not collected during the dataset curation process, which prohibits downstream fairness analyses. For instance, if the patient's race has not been recorded, it is not possible to analyze whether a model trained on that data is fair with respect to that race \cite{Underdiagnosis_2021}. Failure to include sensitive features can generate discrimination and reduce transparency \cite{chen_why_2018}.
  
      \end{itemize}
    \item \textbf{Labels:}
    \begin{itemize}
        \item \textbf{Social bias reflection on labels:}
        Biases in healthcare systems  widely reflect existing biases in society \cite{mamary2018race,sun2020exploring, Vyas_2020}. For instance, race and sex biases exist in COPD underdiagnosis \cite{mamary2018race}, in medical risk score analysis (whereby there exists a higher threshold for Black patients to gain access to clinical resources) \cite{Vyas_2020}, and in the time of diagnosis for cardiovascular disease (whereby female patients are diagnosed much later compared to the male patients with similar conditions) \cite{sun2020exploring}. These biases are reflected in the labels used to train clinical AI systems and, as a result, the model will learn to replicate this bias. 
         \item \textbf{Bias of automatic labeling:}
        Due to the high cost and labour-intensive process of acquiring labels for healthcare data, there has been a shift away from hand-labelled data, towards automatic labelling \cite{bustos2020padchest, irvin_chexpert:_2019,johnson_mimic-cxr:_2019}. For instance, instead of expert-labeled radiology images, natural language processing (NLP) techniques are applied to radiology reports in order to extract labels. This presents concerns as these techniques have shown racial biases, even after they have been trained on clinical notes \cite{Hurtfulwords_2020}. Therefore, using NLP techniques for automatic labeling may sometimes amplify the overall bias of the labels \cite{Underdiagnosis_2021}.  
    
     \end{itemize}

    \item \textbf{Resources:}
    \begin{itemize}
   \item \textbf{Limited computational resources:} Not all centers have enough labeled data or computational resources to train ML models `from scratch' and must use pretrained models for inference or transfer learning. If the original model has been trained on biased (or differently distributed) data, it will unfairly influence the outcome, regardless of the quality of the data at the host center. 
    
\end{itemize}
\end{itemize}
  
\paragraph{Evaluation}
To evaluate the fairness of a model, we need to decide which fairness metric to use and what sensitive attributes to consider in our analysis. 

\begin{itemize}
 
 \item \textbf{Fairness metric(s):} There are many ways to define fairness metrics. For instance, \cite{chouldechova_fair_2016} and \cite{hardt_equality_2016} discussed several fairness criteria and suggested balancing the error rate between different subgroups \cite{Mismeasure, zhang2022improving}. However, it is not always possible to satisfy multiple fairness constraints concurrently \cite{Underdiagnosis_2021}. Jon Kleinberg et al., \cite{kleinberg2016inherent} showed that  three fairness conditions evaluated could not be simultaneously satisfied. As a result, a trade-off between the different notions of fairness is required, or a single fairness metric can be chosen based on domain knowledge and the given clinical application.
 \item \textbf{Sensitive attributes:}
 Sensitive attributes are protected groups that we want to consider when evaluating the fairness of an AI model. Sex and race are two commonly used sensitive attributes \cite{zhang2022improving, CheXclusion_2020, Underdiagnosis_2021, Hurtfulwords_2020}. However, a lack of fairness in an AI system with respect to other sensitive attributes such as age \cite{CheXclusion_2020, Underdiagnosis_2021}, socioeconomic status, \cite{CheXclusion_2020, Underdiagnosis_2021, Hurtfulwords_2020}, and spoken language \cite{Hurtfulwords_2020} are also important to consider.  

\end{itemize}

Defining AI fairness is context- and problem-dependent. For instance, if we build an AI model to support decision making for disease diagnosis with the goal of using it in the clinic, then it is critical to ensure equal opportunity in the model is provided;  i.e., patients from different races should have equal opportunity to be accurately diagnosed \cite{CheXclusion_2020}. However, if an AI model is to be used to triage patients, then ensuring the system does not underdiagnose unhealthy patients of a certain group may be of greater concern compared to the specific disease itself because the patient will lose access to timely care \cite{Underdiagnosis_2021}.

\subsubsection{Interpretability \& Explainability}
\label{interpretability}
In recent years, interpretability has received a lot of interest from the ML community \cite{molnar2020interpretable,tjoa2020survey,markus2021role}. In machine learning, interpretability is defined as the ability to explain the rationale for an ML model's predictions in terms that a human can understand \cite{doshi2017towards} and explainability refers to a detailed understanding of the model's internal representations, {\it a priori} of any decision. After other research in this area \cite{marcinkevivcs2020interpretability}, we use `interpretability' and `explainability' interchangeably.  \\

Interpretability is not a pre-requisite for all AI systems \cite{doshi2017towards, molnar2020interpretable}, including in low-risk environments (in which miscalculations have very limited consequences) and in well-studied problems (which have been tested and validated extensively according to robust MLOps methods). However, interpretability can be crucial in many cases, especially for systems deployed in the healthcare domain \cite{ghassemi2022machine}. The need for interpretability arises from the \textit{incompleteness} of the problem where system results require an accompanying rationale. 

\paragraph{Importance of interpretability}

Interpretability applied to an ML model can be useful for the following reasons:

\begin{itemize}
   
    \item \textbf{Trust:} Interpretability enhances trust when all components are well-explained. This builds an understanding of the decisions made by a model and may help integrate it into the overall workflow.
    
    \item \textbf{Reliability \& robustness:} Interpretability can help in auditing ML models, further increasing model reliability. 
    
    \item \textbf{Privacy \& security:} Interpretability can be used to assess if any private information is leaked from the results. While some researchers claim that interpretability may hinder privacy \cite{sheng2018anatomy,harder2020interpretable,sheng2018anatomy} as the interpretable features may leak sensitive information, others have shown that it can help make the system robust against the adversarial attacks \cite{li2022interpretable,zhang2019interpreting}.
    
    \item \textbf{Fairness:} 
    Interpretability can help in identifying and reducing biases discussed in Sec.~\ref{bias_fairness}. However, the quality of these explanations can differ significantly between subgroups and, as such, it is important to test various explanation models in order to carefully select an equitable model with high overall fidelity \cite{balagopalan2022road}.
    
    \item \textbf{Better understanding and knowledge:} A good interpretation of the model can lead to the identification of the factors that most impact the model. This can also result in a better understanding of the use case itself and enhance  knowledge in that particular area.
    
     \item \textbf{Causality:} Interpretability gives a better understanding of the model decisions and the features and hence can help to identify causal relationships of the features \cite{carvalho2019machine}.
\end{itemize}

\paragraph{Types of approaches for interpretability in ML:}

Many  methods have been developed for better interpretability in ML, such as explainable AI for trees \cite{lundberg2019explainable}, {Tensorflow Lattice}\footnote{\url{https://www.tensorflow.org/lattice}},  DeepLIFT \cite{li2021deep}, InterpretML\cite{nori2019interpretml}, LIME \cite{ribeiro2016should}, and SHAP \cite{lundberg2017unified}. Some of these have been applied to healthcare \cite{abdullah2021review,stiglic2019interpret}. The methods for interpretability are usually categorized as: 
\begin{itemize}

 \item \textbf{Model-based}
    \begin{itemize} 
        \item \textbf{Model-specific:} Model-specific interpretability can only be used for a particular model. Usually, this type of interpretability uses the model's internal structure to analyze the impact of features, for example.
        \item \textbf{Model-agnostic:} Interpretability is not restricted to a specific machine learning model and can be used more generally with several. 
      
    \end{itemize}

\item \textbf{Complexity-based}
    \begin{itemize}
        \item \textbf{Intrinsic:} Relatively simple methods, such as height-bound decision trees, are easier for humans to understand.
        \item \textbf{Post-hoc:} After the model has produced output, interpretation proceeds for more complex methods.
    \end{itemize}

    \item \textbf{Scope-based}
    \begin{itemize}
        \item \textbf{Locally interpretable:} Interprets individual or per-instance predictions  of the model.
         \item \textbf{Globally interpretable:}  Interprets the model's overall prediction set and provides insight into how the model works in general.
    \end{itemize}

    \item \textbf{Methodology-based approach}

        \begin{itemize}
            \item \textbf{Feature-based:} Methods that interpret the models based on the impact of the features on that model. E.g., weight plot, feature selection, etc.
            \item \textbf{Perturbation-based:} Methods that interpret the model by perturbing the settings or features of the model. E.g., LIME \cite{ribeiro2016should}, SHAP \cite{lundberg2017unified} and anchors.
            \item \textbf{Rule-based:} Methods that apply rules on features to identify their impact on the model e.g., BETA, MUSE, and decision trees.
            \item \textbf{Image-based:} Methods where important inputs are shown using images superimposed over the input e.g., saliency maps \cite{adebayo2018sanity}.
        \end{itemize}

\end{itemize}


\paragraph{Interpretability in healthcare} In recent years, interpretability has become common in healthcare \cite{abdullah2021review, meng2022interpretability, rasheed2022explainable}. In particular, Abdullah {\it et al.}~\cite{abdullah2021review} reported that interpretability methods (e.g., decision trees, LIME, SHAP) have been applied to extract insights into different medical conditions including cardiovascular diseases, eye diseases, cancer, influenza, infection, COVID-19, depression, and autism.  Similarly, Meng {\it et al.}~\cite{meng2022interpretability} performed interpretability of deep learning mortality prediction models and fairness analysis on the MIMIC-III dataset \cite{johnson2020mimic}, showing  connections between interpretability methods and fairness metrics.     


\subsubsection{Privacy \& Security}
\label{privacy}


While digitizing healthcare has led to centralized data and improved access for healthcare professionals, it has also increased risks to data security and privacy \cite{newaz2021survey}.
After previous work \cite{abi2019comparative}, {\it privacy} is the individual's ability to control, interact with, and regulate their personal information and {\it security} is a systemic protection of data from leaks or cyber-attacks. 

\paragraph{{Security \& privacy requirements}}
In order to ensure privacy and security, the following requirements should be met  \cite{newaz2021survey}:
\begin{itemize}
    \item \textbf{Authentication:} Strong authentication mechanisms for accessing the system. 
    \item \textbf{Confidentiality:} Access to  data and devices should be restricted to authorized users.
    \item \textbf{Integrity:} Integrity-checking mechanisms should be applied to restrict any modifications to the data or to the system.
    \item \textbf{Non-repudiation:} Logs should be maintained to monitor the system. Access to those logs should be restricted and avoid any tampering. 
    \item \textbf{Availability:} Quick, easy, and fault-tolerant availability should be ensured at all times. 
    \item \textbf{Anonymity:} Anonymity of the device, data, and communication should be guaranteed. 
    \item \textbf{Device unlinkability:} An unauthorized person should not be able to establish a connection between data and the sender.
    \item \textbf{Auditability and accountability:} It should be possible to trace back the recording time, recording person, and origins of the data to validate its authenticity. 
    
\end{itemize}

\paragraph{Types of threats}
Violation of privacy \& security can occur either due to human error (unintentional or non-malicious) or an adversarial attack (intentional or malicious).
\begin{enumerate}
\item \textbf{Human error:}
Human error can cause data leakage through the carelessness or incompetence of authorized individuals. Most of the literature in this context \cite{liginlal2009significant, evans2019heart} divides human error into two types: 
\begin{enumerate}
    \item \textbf{Slip:} the wrong execution of correct, intended actions; e.g., incorrect data entry, forgetting to secure the data, giving access of information to unauthorized persons using the wrong email address.  
    \item \textbf{Mistake:} the right execution of incorrect, unintended actions; e.g., collecting data that is not required, using the same password for different systems to avoid password recovery, giving access of information to unauthorized persons assuming they can have access. 
\end{enumerate}

   While people dealing with data should be trained to avoid such negligence, some researchers have suggested policies, frameworks, and strategies such as \textit{error avoidance}, \textit{error interception}, or \textit{error correction} to prevent or mitigate these issues \cite{liginlal2009significant, evans2019heart}.
   
    \item \textbf{Adversarial attacks:}
A primary risk for any digital data or system is from  adversarial attackers \cite{gupta2021responsible} who can damage, pollute, or leak information from the system. An adversarial attacker can attack in many ways; e.g., they can be remote or physically present, they can access the system through a third-party device, or they can be personified as a patient \cite{newaz2021survey}. The most common types of attacks are listed below.

\begin{itemize}
    \item \textbf{Hardware or software attack:} Modifying the hardware or software to use it for malicious purposes.
    \item \textbf{System unavailability:} Making the device or data unavailable.
    \item \textbf{Communication attack:} Interrupting the communication or forcing a device to communicate with unauthorized external devices.
    \item \textbf{Data sniffing:} Illegally capturing the communication to get sensitive information.
    \item \textbf{Data modification:} Maliciously modifying data.
    \item \textbf{Information leakage:} Retrieving sensitive information from the system.
\end{itemize}

\end{enumerate}
 \paragraph{Healthcare components and security \& privacy} Extra care needs to be taken to protect healthcare data \cite{abouelmehdi2017big}. Components  \cite{oh2021comprehensive} include:
 \begin{itemize}
     \item \textbf{Electronic health data:} This data can be leaked due to human mistakes or malicious attacks, which can result in tampering or misuse of data. In order to overcome such risks, measures such as access control, cryptography, anonymization, blockchain, steganography, or watermarking can be used.
     \item \textbf{Medical devices:} Medical devices such as smartwatches and sensors are also another source of information that can be attacked.  Secure hardware and software, authentication and cryptography can be used to avoid such problems. 
     \item \textbf{Medical network:} Data shared across medical professionals and organizations through a networks may be susceptible to eavesdropping, spoofing, impersonating, and unavailability attacks. These threats can be reduced by applying encryption, authentication, access control, and compressed sensing.
     \item \textbf{Cloud storage:} Cloud computing is becoming widely adopted in healthcare. However, like any system, it is also prone to unavailability, data breaches, network attacks, and malicious access. Similar to those above, threats to cloud services can be avoided through authentication, cryptography, and decoying (i.e., a method to make an attacker erroneously believe that they have acquired useful information). 
 \end{itemize}

\paragraph{Healthcare privacy \& security laws}  Due to the sensitivity of healthcare data and communication, many countries have introduced laws and regulations such as the Personal Information Protection and Electronic Documents Act (PIPEDA) in Canada, the Health Insurance Portability and Accountability
Act (HIPPA) in the USA, and the Data Protection Directive in the EU \cite{xiang2021privacy}. These acts mainly aim at protecting patient data from being shared or used without their consent but while allowing them to access to their own data.

\paragraph{Attacks on ML pipeline }
Any ML model that learns from data can also leak information about the data, even if it is generalized well; e.g., using membership inference (i.e., determining if a particular instance was used to train the model) \cite{melis2019exploiting, hu2022membership} or using  property inference (i.e., inferring properties of the training dataset from a given model) \cite{melis2019exploiting, parisot2021property}.
Adversarial attacks in the context of the MLOps pipeline can occur in the following phases \cite{gupta2021responsible}:

\begin{itemize}
    \item \textbf{Data collection phase:} At this phase, a \textit{poisoning attack}  results in modified or polluted data, impacting the training of the model and lowering performance on unmodified data.
    
    \item \textbf{Modelling phase:} Here, the \textit{Trojan AI attack} can modify a model to provide an incorrect response for specific trigger instances \cite{wang2022survey} by changing the model architecture and parameters. Since  it is now common to use pre-trained models, these models can be modified or replaced by attackers. 

    \item \textbf{Production and deployment phases:} At these phases, both \textit{Trojan AI attacks} and \textit{evasion attacks} can occur. Evasion attacks consist, e.g., of modifying test data to have them misclassified \cite{pitropakis2019taxonomy}.

\end{itemize}

\subsubsection{Reliability, robustness and resilience}
\label{3r}

A trustworthy MLOps system should be reliable, robust, and resilient. These terms are defined as follows \cite{zissis2019r3}:
\begin{itemize}
    \item \textbf{Reliability:} The system performs in a satisfactory manner under specific, unaltered operating conditions.
    \item \textbf{Robustness:} The system performs in a satisfactory manner despite changes in operating conditions, e.g., data shift.
    \item\textbf{Resilience:} The system performs in a satisfactory manner despite a major disruption in operating conditions; e.g., adversarial attacks.
\end{itemize}

These aspects have been studied in the healthcare domain \cite{mincu2022developing,qayyum2020secure} and different approaches such as interpretability, security, privacy, and methods to deal with data shift (discussed in Sections \ref{interpretability} and \ref{privacy}) have been suggested.\\

\textbf{Trade-off between accuracy and trustworthiness:} In Section \ref{defresAI}, we discussed different important components of trustworthy AI that should be considered while designing an ML system; however,  literature shows that there can be a trade-off between accuracy, interpretability, and robustness  \cite{rasheed2022explainable,tsipras2018robustness}.  While a main reason for the trade-off is that robust models learn a different feature representation that may decrease accuracy, it is better perceived by humans \cite{tsipras2018robustness}. 




\subsection{Incorporating Responsibility and Trust into MLHOps}
\label{resmlhops}
In recent years, responsible and trustworthy AI have gained a lot of attention in general as well as for healthcare due to its implications on society \cite{rasheed2022explainable}.  
There are several definitions of trustworthiness \cite{rasheed2022explainable}, and they are related to making the system robust, unbiased, generalizable, reproducible, transparent, explainable, and secure. However,  the lack of standardized practices for applying, explaining, and evaluating trustworthiness in AI for healthcare makes this very challenging \cite{rasheed2022explainable}. In this section, we discuss how we can incorporate all these qualities at each step of the pipeline.

\subsubsection{Data}
The process of a responsible and trustworthy MLOps pipeline starts with  data collection
and preparation. The impact of biased or polluted data propagates through all the subsequent steps of the pipeline \cite{fuchs2018dangers}. This can be even more important and challenging in the healthcare domain due to the privacy and sensitivity of the data \cite{awotunde2021privacy}. If compromised, this information can be tempered or misused in various ways (e.g., identity theft, information sold to a third party) and  introduce bias in the healthcare system. Such challenges can also cause economic harm (such as job loss), psychological harm (e.g., causing embarrassment due to a medical issue), and social isolation (e.g., due to a serious illness such as HIV) \cite{nass2009value, abouelmehdi2018big}. It can also impact  ML model performance and trustworthiness \cite{chen2021ethical}.

    \paragraph{Data collection} In healthcare, data can be acquired through multiple sources \cite{ullah2021secure}, which increases the chance of the data being polluted by bias. Bias can concern, for example, race\cite{yala2019fairness}, gender, sexual orientation, gender identity, and disability. Bias in healthcare data can be mitigated against by increasing diversity in data, e.g., by including underrepresented minorities (URMs), which can lead to better outcomes \cite{marcelin2019impact}. Debiasing during data collection can include:
    \begin{enumerate}
    \item \textbf{Identifying \& acknowledging potential real-world biases:}
    Bias in healthcare is introduced long before the data collection stage. Although increasingly less common in many countries\footnote{{\url{https://applymd.utoronto.ca/admission-stats}}\label{UofT}}, bias can still occur in medical school admission,  job interviews, patient care, disease identification, research samples, and case studies. Such biases lead to the dominance of people from certain communities  \cite{marcelin2019impact} or in-group vs. out-group bias \cite{graham2021ingroup}, which can result in stereotyped and biased data generation and hence biased data collection.
   
    Bias can be unconscious or conscious \cite{marcelin2019impact,fitzgerald2017implicit}. Unconscious bias stems from implicit or unintentional associations outside conscious awareness resulting from stereotypical perceptions and experiences. On the other hand, conscious bias is explicit and intentional and has resulted in abuse and criminal acts in healthcare; e.g., the Tuskegee study of untreated Syphilis in black men demonstrated intentional racism  \cite{francis2001medical}. Both conscious and unconscious biases damage the validity of the data. Since conscious bias is relatively more visible, it is openly discouraged not only in healthcare but also in all areas of society. However, unconscious bias is more subtle and not as easy to identify. In most cases, unconscious bias is not even known to the person suffering from it. 

    Different surveys, tests, and  studies have found the following types of biases (conscious or unconscious) common in healthcare \cite{marcelin2019impact}:
    \begin{enumerate}
        \item\textbf{Racial bias} e.g., Black, Hispanic, and Native American physicians are underrepresented \cite{osseo2018minority}. According to one study, white males from the upper classes are preferred by the admission committees \cite{capers2017implicit} (although some other sources suggest the opposite$^{28}$).
        \item \textbf{Gender bias:} e.g., professional women in healthcare being less likely to be invited to give talks \cite{mehta2018speaker}, to be introduced using professional titles \cite{files2017speaker}, to experience harassment or exclusion, to receive insufficient support at work or negative comparisons with male colleagues, and to be perceived as weak \& less competitive \cite{lim2021unspoken,templeton2019gender}.

        \item \textbf{Gender minority bias} e.g., LGBTQ people receive lower quality healthcare \cite{rosenkrantz2017health} and faced challenges to get jobs in healthcare \cite{sanchez2015lgbt}. 
        
        \item \textbf{Disability bias} e.g., people with disabilities  receive limited accessibility supports to all facilities and have to work harder to be feel validated or recognized \cite{meeks2018removing}.
    \end{enumerate}

    Various tests identify the existence of unconscious bias, such as the Implicit Association Test (IAT), and have been reported to be useful. For example, Race IAT results detected unintentional bias in $75\%$ of the population taking the test   \cite{banaji2013blindspot}. While debate continues regarding the degree of usefulness of these tests \cite{blanton2009strong}, they may still capture some subtle human behaviours. Some other assessment tools (e.g., Diversity Engagement Survey (DES) \cite{person2015measuring}) have also been built for successfully measuring inclusion and diversity in medical institutes.
    
    According to Marcelin et al.~\cite{marcelin2019impact}, the following measures can help in reducing unintentional bias:
    
    \begin{enumerate}
        \item Using IAT to identify potential biases in  admissions or hiring committee members in advance.
        \item Promoting  equity, diversity, inclusion, and accessibility (EDIA) in  teams. Including more people from underrepresented minorities (URM) in the healthcare profession, especially in admissions and hiring committees.
        \item Conducting and analyzing surveys to keep track of the challenges faced by URM individuals due to the biased perception of them.  
        \item Training to highlight the existence and need for mitigation of bias.
        \item Self-monitoring bias can be another way to incorporate inclusion and diversity.
    \end{enumerate}

    \item \textbf{Debiasing during data collection and annotation:}\\
    In addition to human factors, we can take steps to improve the data collection process itself. In this regard, the following measures can be taken \cite{liu2021trustworthy}: 
    
     \begin{enumerate}
         
         \item \textbf{Investigating the exclusion:} In dataset creation, an important step is to carefully investigate which patients are included in the dataset. An exclusion criterion in dataset creation may be conscious and clinically motivated, but there are many unintentional exclusion criteria that are not very well visible and enforce biases. For instance, a dataset that is gathered in a research hospital that does not routinely provide standard diagnostic and treatment services and select the patients only because they have an illness being studied by the Institutes have a different type of patients compared to clinical hospitals that do not have these limitations  \cite{Underdiagnosis_2021}. Alternatively, whether the service delivered to the patient is free or covered by insurance would change the distribution of the patients and infect biases into the resulting AI model \cite{CheXclusion_2020}. 

         \item \textbf{Annotation with explanation:} Adding justification for choosing the label by the human annotators not only helps them identify their own unconscious biases but also can help in setting standards for unbiased annotations and avoid any automatic association and stereotyping (e.g., high prevalence HIV in gay men led to under-diagnosis of this disease in women and children \cite{marcelin2019impact}. Moreover, these explanations can be a good resource for training explainable AI models \cite{wiegreffe2021teach}.
         
         \item \textbf{Data provenance:} This involves tracking data lineage through the data source, dependencies, and data collection process. Healthcare data can come from multiple sources which increases the chances of it being biased \cite{chaudhary2022taxonomy}. Data provenance improves data quality, integrity, audibility, and transparency \cite{xu2018application}. Different tools for data provenance are available including  \textit{Fast Healthcare Interoperability Resources (FHIR)} \cite{saripalle2019fast} and \textit{Atmolytics}\cite{xu2018application}. \cite{margheri2020decentralised} 

         \item \textbf{Data security \& privacy during data collection:} Smart healthcare technologies have become a common practice \cite{chaudhary2022taxonomy}. A wide variety of smart devices is available, including wearable devices (e.g., smartwatches, skin-based sensors), body area networks (e.g., EEG sensors, blood pressure sensors),  tele-healthcare (e.g., tele-monitoring, tele-treatment), digital healthcare systems (e.g., electronic health records (EHR), electronic medical records (EMR)), and health analytics (e.g., medical big-data). While the digitization of healthcare has improved access to medical facilities, it has increased the risk of data leakage and malicious attacks. Extra care should be taken while designing an MLOps pipeline to avoid privacy and security risks, as it can lead to serious life-threatening consequences. Other issues include the number of people involved in using the data and proper storage for high volumes of data. Chaudhry et al.~\cite{chaudhary2022taxonomy} proposed an AI-based framework using 6G-networks for secure data exchange in digital healthcare devices. In the past decade, the blockchain has also emerged as a way of ensuring data privacy and security. Blockchain is a distributed database with unique characteristics such as immutability, decentralization, and transparency. This is especially relevant in healthcare because of  security and privacy issues \cite{haleem2021blockchain, yaqoob2019use,ng2021blockchain}. Using blockchain can help in more efficient and secure management of patient's health records, transparency, identification of false content, patient monitoring, and maintaining financial statements  \cite{haleem2021blockchain}.  

         \item \textbf{Data-sheet:} Often, creating a dataset that represents the full diversity of a population is not feasible, especially for very multi-cultural societies.  Additionally, the prevalence of diseases among different sub-populations may be different \cite{Underdiagnosis_2021}. If it is not possible to build  an ideal dataset with the above specifications, the data needs to be delivered by a data-sheet. The data-sheet is meta-data that helps to analyze and specify the characteristics of the data, clearly explain exclusion and inclusion criteria, detail demographic features of the patients, and statistics of  the data distribution over sub-populations, labels and features. 

     \end{enumerate}

\end{enumerate}  
\paragraph{Data pre-processing}
    \begin{enumerate}
    
        \item \textbf{Data quality assurance:}  Sendak et al.~\cite{sendak2022development} argued that clinical researchers choose data for research very carefully but the machine learning community in healthcare does not follow this practice. To overcome this gap, they suggest that data points are identified by the clinicians and extracted into a project-specific data store. After this, a three-step framework is applied: (1) use different measures for data pre-processing to ensure the correctness of all data elements (e.g, converting each lab measurement to the same unit), (2)  ensure completeness, conformance, plausibility, and possible data shifts, and (3) adjudicate the data with the clinicians.
        
        \item \textbf{Data anonymization:}  Due to the sensitivity of healthcare data preparation, data anonymization should minimize the chances of it being de-anonymized.  Olatunji et al.~\cite{olatunji2022review} provide a detailed overview of data anonymization models and techniques in healthcare such as k-anonymity, k-map, l-diversity, t-closeness, $\delta$-disclosure privacy, $\beta$-likeness, $\delta$-presence, and ($\epsilon$, $\delta$)-differential privacy. To avoid data leakage, many tools for data anonymization and its evaluation tools \cite{vovk2021evaluation} such as SecGraph \cite{ji2015secgraph},  ARX- tool for anonymizing biomedical data \cite{prasser2014arx}, Amnesia\footnote{\url{https://www.openaire.eu/item/amnesia-data-anonymization-made-easy}} \cite{tomas2022data}, PySyft \cite{ryffel2018generic},  Synthea \cite{walonoski2018synthea} and Anonimatron\footnote{\url{https://realrolfje.github.io/anonimatron/}} (open-source data anonymization tool written in Java) can be incorporated in the MLHOps pipeline.


        \item \textbf{Removing subgroups indicators}. Changing the race of the patients can have a dramatic impact on the outcome of an algorithm that is designed to fill a prompt \cite{Hurtfulwords_2020}.  Therefore, the existence of race attributes in the text can decrease the fairness of the model dramatically. In some specific problems, removing  subgroup indicators such as the sex of a job candidate from their application has shown to have minimal influence on classifier accuracy while improving the fairness \cite{de2019bias}. This method is applicable mostly in text-based data where sensitive attributes are easily removable. As a preprocessing  step, one can estimate the effect of keeping or removing such sensitive attributes on the overall accuracy and fairness of a developed model. At the same time, it is not always possible to remove the sensitive attributes from the data. For example, AI models can predict patient race from medical images, but it is not yet clear {\it how} they can do it \cite{racedetection}. In one study \cite{racedetection}, researchers did not provide the patient race during model training, but they also could not find a particular patch or region in the data for which AI failed to detect race by removing that part. 

        \item \textbf{Differential privacy:}
         Differential privacy \cite{dankar2013practicing}  aims to provide information about inherent groups while withholding the information about the individuals. Many algorithms and tools have been developed for this, including CapC\cite{choquette2021capc} and PySyft \cite{ryffel2018generic}.
    \end{enumerate}



\subsubsection{Methodology}

The following sections overview the steps to put these concepts into practice.
   
    \paragraph{Algorithmic fairness} Algorithmic fairness \cite{mitchell2021algorithmic, xu2022algorithmic, galhotra2022causal} attempts to ensure the unbiased output
across the available classes.
    Here, we discuss how we can overcome this challenge at different stages of model training \cite{mitchell2021algorithmic, xu2022algorithmic}.
    \begin{enumerate}
\item \textbf{Pre-processing}
\begin{itemize}
    \item \textit{Choice of sampling \& data augmentation:} Making sure that the dataset is balanced (having approximately an equal number of instances from each class) and all the classes get equal representation in the dataset using simple under- or over-sampling methods \cite{xu2022algorithmic}. This can also be done by data augmentation \cite{mhasawade2021machine,fryer2022flexible} to improve the counterfactual fairness by counterfactual text generation and using it to augment data. Augmentation methods include  \textit{Synthetic Minority Oversampling Technique} (SMOTE) \cite{chawla2002smote} and Adaptive Synthetic Sampling (ADASYN) \cite{he2008adasyn}. Since synthetic samples may not be universally beneficial for the healthcare domain, acquiring more data and undersampling may be the best strategy \cite{xu2022algorithmic}.
    
    \item \textit{Causal fairness using data pre-processing:} 
    Causal fairness is achieved by reducing the impact of protected or sensitive attributes (e.g., race and gender) on predicted variables and different methods have been developed to accomplish this \cite{galhotra2022causal,xu2019achieving}. Kamiran {\em et al.} \cite{kamiran2009classifying} proposed ``massaging the data'' before using traditional classification algorithms.
   \item \textit{Re-weighing:} In a pre-processing approach, one may re-weight the training dataset samples or remove features with high correlation to sensitive attributes as well as the sensitive attribute itself \cite{kamiran2012data}, learning representations that are relatively invariant to sensitive attribute \cite{louizos2015variational}. One might also adjust representation rates of protected groups and achieve target fairness metrics \cite{entropy2020mitigate}, or utilize optimization to learn a data transformation that reduce discrimination \cite{ Discrimination2017Prevention}.

\end{itemize}

   \item \textbf{In-processing}\\

\begin{itemize}
    \item \textit{Adversarial learning:} It is also possible to enforce fairness during model training, using adversarial debiasing \cite{Mitigating2018Association, Conditional2021Adversarial, Balanced2019Datasets}. Adversarial learning refers to the methods designed to intentionally confound ML models during training, through deceptive or misleading inputs, to make those models more robust. This technique has been used in healthcare to create robust models \cite{kaviani2022adversarial}, and for bias mitigation,  by intentionally inputting biased examples  \cite{li2021estimating,pfohl2019creating}.
     \item \textit{Prejudice remover:} Another important aspect is prejudice injected into the features \cite{kamishima2012fairness}. Prejudice can be (a) \textit{Direct prejudice}: using a protected attribute as a prediction variable, (b) \textit{Indirect prejudice}: statistical dependence between protected attributes and prediction variables, and (c) \textit{Latent prejudice}: statistical dependence between protected attributes and non-protected attributes. Kamishaima {\em et al.} \cite{kamishima2012fairness} proposed a method to remove prejudice using regularization. Similarly, Grgic {\em et al.} \cite{grgic2016case} introduced a method using constraints for classifier optimization objectives to remove prejudice.


    \item \textit{Enforcing fairness in the model training:}
Fairness can also be enforced by making changes to the model through constraint optimization \cite{FairALM2020Augmented},  modifying loss functions to penalize deviation from the general population for subpopulations \cite{Pfohl_2021}, regularizing loss function to minimize mutual information between feature embedding and bias \cite{Learning2018Not}, or adding regularizer to identify and treat latent discriminating features  \cite{kamishima2012fairness}. 

\item \textit{Up-weighing:} It is possible to improve the outcome on worst case group by up weighting the groups with the largest loss \cite{ zhang2022improving, sagawa2019distributionally, Blind2021Robustness}. However, all these methods need awareness about the membership of the instance to sensitive attributes. There are also group un-aware methods where they try to weights each sample with an adversary that tries to maximize the weighted loss \cite{Adversarially2020Neural}, or trains an additional classier that up-weights samples classified incorrectly in the last training step \cite{Twice2021PMLR}.     

\end{itemize}
\item \textbf{Post-processing:}
The post-processing fairness mitigation approaches may target post-hoc calibration of model predictions. This method has sown impact in bias mitigation in both non-healthcare \cite{hardt_equality_2016, Fairness2017inc} and healthcare \cite{Multiaccuracy2019black} applications.  
    \end{enumerate}

There are some software tools and
libraries for algorithmic fairness check, listed in \cite{xu2022algorithmic}, which can be used by developer and end user to evaluate the fairness of the AI model outcomes.

\subsubsection{Development \& evaluation}

At this stage, the ML system is evaluated to make sure its trustworthiness, which includes evaluating the evaluation methods \cite{rasheed2022explainable,antoniou2021health}.



     



\paragraph{Model interpretability \& explainability}

At this stage, model evaluation can be done through interpretability and explainability methods to mitigate any potential issues such as possible anomalies in the data or the model. However, it should be noted that the methods perform interpretability and explainability should also be evaluated carefully before relying on them, which can be performed using different methods such as human evaluation \cite{marcinkevivcs2020interpretability,balagopalan2022road}.

\section{Concluding remarks}


Machine learning (ML) has been applied to many clinically-relevant tasks and many relevant datasets in the research domain but, to fully realize the promise of ML in healthcare, practical considerations that are not typically necessary or even common in the research community must be carefully designed and adhered to. We have provided a deep survey into a breadth of these ML considerations, including infrastructure, human resources, data sources, model deployment, monitoring and updating, bias, interpretability, privacy and security.\\

As there are an increasing number of AI systems being deployed into medical practice, it is important to standardize and to specify specific engineering pipelines for medical AI development and deployment, a process we term MLHOps. To this end, we have outlined the key steps that should be put into practice by multidisciplinary teams at the cutting-edge of AI in healthcare to ensure the responsible deployment of clinical AI systems.






\section{Appendix}

\begin{table}[ht]
\begin{center}
\begin{tabular}{ |p{3cm}||p{9cm}| }
 \hline
 Name of tool &  Description \\
 \hline
 \texttt{MIMIC-Extract} & Pipeline to transform data from MIMIC-III into DataFrames that are directly usable for ML modelling \\
 \texttt{Clairvoyance}  &  End-to-End AutoML Pipeline for Medical Time Series \\
 \texttt{Pyhealth} &  A python library for health predictive models \\
 \texttt{ROMOP} &  R package to easily interface with OMOP-formatted EHR data \\
 \texttt{ATLAS} &  Research tool to conduct scientific analyses on data available in OMOP format \\
 \texttt{FIDDLE} & Preprocessing pipeline that transforms structured EHR data into feature vectors for clinical use cases \\
 \texttt{hi-ml} & Toolbox for deep learning for medical imaging and Azure integration \\
 \texttt{MedPerf} & An open benchmarking platform for medical artificial intelligence using Federated Evaluation. \\
 \texttt{MONAI} & AI Toolkit for Healthcare Imaging \\
 \texttt{TorchXRayVision} & A library of chest X-ray datasets and models \\
 \texttt{Leaf} & Clinical Data Explorer \\
 \hline
\end{tabular}
\end{center}
\label{mlhops_tools}
\caption{List of open-source tools available on Github that can be used for ML system development specific to health.}
\end{table}

\begin{table}
\centering
\caption{Key Roles in an MLOps Team}
\label{tab:mlops roles}
\begin{tabular}{p{3.5cm}|p{4.5cm}|p{6.5cm}}
\hlineB{4}
{\textbf{Role}} & 
{\textbf{Alternatively}} &
{\textbf{Description}}  \\ 
\hlineB{4}
Domain Expert & \begin{tabular}{@{\labelitemi\hspace{\dimexpr\labelsep+0.5\tabcolsep}}l@{}}\textit{Business Translator}\\\textit{Business Stakeholder}\\\textit{PO/Manager}\\\end{tabular} & An instrumental role in any phase of the MLOps process where a deeper understanding of the data and the domain is required. \\ 
\hline
Solution Architect                 & \begin{tabular}{@{\labelitemi\hspace{\dimexpr\labelsep+0.5\tabcolsep}}l@{}}\textit{IT Architect}\\\textit{ML Architect}\end{tabular}  & Unifying the work of data scientists, data engineers, and software developers through developing strategies for MLOps processes, defining the project lifecycle, and identifying the best tools and assemble the team of engineers and developers to work on projects.   \\ 
\hline
Data Scientist                    & \begin{tabular}{@{\labelitemi\hspace{\dimexpr\labelsep+0.5\tabcolsep}}l@{}}\textit{ML Specialist}\\\textit{ML~Developer}\end{tabular}  &  A central player in any MLOps team responsible for creating the data and ML model pipelines. The pipelines include analysing and processing the data as well as building and testing the ML models. \\ 
\hline
Data Engineer                     & \begin{tabular}{@{\labelitemi\hspace{\dimexpr\labelsep+0.5\tabcolsep}}l@{}}\textit{DataOps Engineer}\\\textit{Data Analyst}\end{tabular}  & Working in coordination with product manager and domain expert to uncover insights from data through data ingestion pipelines.                                            \\ 
\hline
Software Developer                 & \begin{tabular}{@{\labelitemi\hspace{\dimexpr\labelsep+0.5\tabcolsep}}l@{}}\textcolor[rgb]{0.125,0.129,0.141}{\textit{Full-stack engineer}}\end{tabular}  &   Focusing on the productionizing of ML models and the supporting infrastructure based on the ML architect's blueprints. They standardize the
code for compatibility and re-usability \\ 
\hline
DevOps Engineer                   &  \begin{tabular}{@{\labelitemi\hspace{\dimexpr\labelsep+0.5\tabcolsep}}l@{}}\textit{CI/CD Engineer}\\\end{tabular}     &   Facilitating access to the specialized tools and high performance computing infrastructure, enabling transition from development to deployment and monitoring, and automating ML lifecycle.
\\ 
\hline
ML Engineer                       & \begin{tabular}{@{\labelitemi\hspace{\dimexpr\labelsep+0.5\tabcolsep}}l@{}}\textit{MLOps Engineer}\end{tabular}  &  Highly skilled programmers supporting designing and deploying ML models in close collaboration with Data Scientists and DevOps Engineers. \\
\hlineB{4}
\end{tabular}
\end{table}


\begin{table}[ht]
\vspace{-2.5cm}
\begin{center}
\begin{tabular}{p{4.5cm} |p{11cm} } 
 \hline
\textbf{Tool} &\textbf{Description} \\
 \hline
 FairMLHealth\footnote{\url{https://github.com/KenSciResearch/fairMLHealth}} & Tools and tutorials for variation analysis in healthcare machine learning. \\
 
AIF360 \cite{bird2020fairlearn} & An open-source library containing techniques developed by the research community to help detect and mitigate bias in machine learning models throughout the AI application lifecycle.  \\

Fairlearn\footnote{\url{https://fairlearn.org/}} & An open-source, community-driven project to help data scientists improve the fairness of AI systems. \\ 

Fairness-comparison\footnote{\url{https://github.com/algofairness/fairness-comparison}} & Benchmark fairness-aware machine learning techniques. \\


Fairness Indicators\footnote{\url{https://www.tensorflow.org/responsible_ai/fairness_indicators/tutorials/Fairness_Indicators_Example_Colab}} & Fairness Indicators is a suite of tools built on top of TensorFlow Model Analysis (TFMA) that enable regular evaluation of fairness metrics in product pipelines.  \\

ML-fairness-gym\footnote{\url{https://ai.googleblog.com/2020/02/ml-fairness-gym-tool-for-exploring-long.html}} & A tool for exploring long-term impacts of ML systems.\\  

themis-ml \cite{bantilan2018themis} & An open source machine learning library that implements several fairness-aware methods that comply with the sklearn API.\\

FairML \cite{adebayo2016fairml} &  ToolBox for diagnosing bias in predictive modelling. \\ 

Black Box Auditing \cite{adler2018auditing} &  A toolkit for auditing ML model deviations.  \\

What-If Tool\footnote{\url{https://pair-code.github.io/what-if-tool/}} & Visually probe the behaviour of trained machine learning models, with minimal coding.\\

Aequitas\footnote{\url{http://www.datasciencepublicpolicy.org/our-work/tools-guides/aequitas/}} & An open source bias audit toolkit for machine learning developers, analysts, and  policymakers to audit machine learning models for discrimination and bias, and make informed and equitable decisions around developing and deploying predictive risk-assessment tools.  \\

DECAF \cite{van2021decaf} & A fair synthetic data generator for tabular data utilizing GANs and causal models. \\

REPAIR \cite{li2019repair} & A dataset resampling algorithm to reduce representation bias by reweighting.  \\

CERTIFAI \cite{sharma2020certifai} & Evaluates AI models for robustness, fairness, and explainability, and allows users to compare different models or model versions for these qualities.\\


FairSight \cite{ahn2019fairsight} & A fair decision making pipeline to assist decision makers track fairness throughout a model.  \\

Adv-Demog-Text \cite{elazar2018adversarial} & An adversarial network demographic attributes remover from text data.  \\

GN-GloVe \cite{zhao2018learning} & A framework for generating gender neutral word embeddings. \\

\small{Tensorflow Constrained Optimization} \footnote{\url{https://github.com/google-research/tensorflow_constrained_optimization}}  &  A library for optimizing inequality-constrained problems using rate helpers. \\

Responsibly\footnote{\url{https://github.com/ResponsiblyAI/responsibly}} \cite{cotter2019training}  & Toolkit for auditing and mitigating bias and fairness of ML systems.  \\

\small{Dataset-Nutrition-Label} \cite{holland2018dataset} & The Data Nutrition Project aims to create a standard label for interrogating datasets. \\
 \hline
\end{tabular}
\end{center}
\label{fairness_tools}
\caption{List of open-source tools available on Github that can be used for ML Monitoring and Updating specific to health.}
\end{table}

\bibliography{mlops_bib}
\bibliographystyle{plain}

\end{document}